\documentclass[10pt,twocolumn,letterpaper]{article}

\usepackage[pagenumbers]{iccv} 

\usepackage{bm}
\usepackage{colortbl}
\usepackage{multirow}

\definecolor{mygray}{gray}{.85}
\newcommand{\abs}[1]{\left\vert#1\right\vert}
\newcommand{\norm}[1]{\left\Vert#1\right\Vert}

\newcommand{\tabincell}[2]{\begin{tabular}{@{}#1@{}}#2\end{tabular}}

\definecolor{iccvblue}{rgb}{0.21,0.49,0.74}
\usepackage[pagebackref,breaklinks,colorlinks,allcolors=iccvblue]{hyperref}

\def\paperID{***} 
\def\confName{ICCV}
\def\confYear{2025}

\title{Learning Normals of Noisy Points by Local Gradient-Aware Surface Filtering}

\author{
Qing Li$^1$
\quad
Huifang Feng$^2$\thanks{Corresponding author}
\quad
Xun Gong$^1$
\quad
Yu-Shen Liu$^3$\\
$^1$ Southwest Jiaotong University, Chengdu, China\\
$^2$ Xihua University, Chengdu, China
\quad
$^3$ Tsinghua University, Beijing, China\\
{\tt\small qingli@swjtu.edu.cn} \quad {\tt\small fhf@xhu.edu.cn} \quad {\tt\small xgong@swjtu.edu.cn} \quad {\tt\small liuyushen@tsinghua.edu.cn}
}

\begin{document}
\maketitle

\begin{abstract}

    Estimating normals for noisy point clouds is a persistent challenge in 3D geometry processing, particularly for end-to-end oriented normal estimation.
    Existing methods generally address relatively clean data and rely on supervised priors to fit local surfaces within specific neighborhoods.
    In this paper, we propose a novel approach for learning normals from noisy point clouds through local gradient-aware surface filtering.
    Our method projects noisy points onto the underlying surface by utilizing normals and distances derived from an implicit function constrained by local gradients.
    We start by introducing a distance measurement operator for global surface fitting on noisy data, which integrates projected distances along normals.
    Following this, we develop an implicit field-based filtering approach for surface point construction, adding projection constraints on these points during filtering.
    To address issues of over-smoothing and gradient degradation, we further incorporate local gradient consistency constraints, as well as local gradient orientation and aggregation.
    Comprehensive experiments on normal estimation, surface reconstruction, and point cloud denoising demonstrate the state-of-the-art performance of our method.
    The source code and trained models are available at \textcolor{red}{\href{https://github.com/LeoQLi/LGSF}{https://github.com/LeoQLi/LGSF}}.

\end{abstract}

\section{Introduction}   \label{sec:intro}

Point clouds are indispensable in 3D computer vision and play a foundational role in applications such as virtual reality, autonomous driving, and robotic perception.
Surface normal estimation, as a fundamental task in 3D point cloud analysis, is critical for understanding object geometry and supporting downstream tasks like surface reconstruction~\cite{kazhdan2006poisson,kazhdan2013screened} and segmentation~\cite{khaloo2017robust}.
However, the presence of noise in point clouds remains a major challenge, often distorting geometry and impeding precise normal estimation.
Traditional methods~\cite{ben2020deepfit,zhu2021adafit,li2022hsurf,li2023NGLO,li2023shsnet,li2024shsnet-pami} that rely on supervised learning require extensive labeled data and struggle with noisy, unstructured data, making it challenging to obtain reliable normals from corrupted point clouds.

To address these limitations, we propose a novel approach that leverages local gradient-aware surface filtering for estimating oriented normals in noisy point clouds.
Inspired by recent advancements in neural implicit representations, we adopt techniques from implicit function learning to bridge the gap between raw point clouds captured by 3D sensors and the smooth, continuous surfaces required for inferring accurate normals.
Unlike existing methods~\cite{atzmon2020sal,ma2020neural,zhou2024cap,li2025implicit} that focus solely on individual point constraints, often resulting in over-smoothing or gradient degradation, our method can recover high-quality 3D geometry from noisy observations by introducing specialized loss functions with local gradient constraints.

To learn the surface representations, we introduce a distance measurement operator that enables global surface fitting from noisy data by incorporating projected distances along normals.
We propose implicit field-based filtering to project points onto the underlying surface based on normals and distances derived from an implicit function, which is defined through signed distance fields and local gradient constraints.
To properly guide the projection during the filtering process, we incorporate the constraints of local gradient consistency, orientation and aggregation to mine the geometric information in noisy point cloud data.
The surface filtering effectively reduces noise while maintaining the shape's intricate geometries, allowing us to achieve a refined and noise-resilient surface representation.
To demonstrate the effectiveness of our method, we evaluate it on three key tasks in point cloud processing: normal estimation, surface reconstruction, and point cloud denoising.
Experimental results show that our approach significantly improves performance on noisy data, highlighting its robustness and suitability for practical 3D vision applications.
In summary, our main contributions include:
\begin{itemize}[leftmargin=*]
\setlength{\itemsep}{0pt}
\setlength{\parsep}{0pt}
\setlength{\parskip}{0pt}

\item We propose a new paradigm for surface fitting from noisy point clouds by conducting filtering using normals and distances derived from an implicit function.

\item We introduce the local gradient consistency constraints, local gradient orientation and aggregation to enhance the surface filtering for learning normals.

\item We report the state-of-the-art performance of our method across three tasks in point cloud processing.

\end{itemize}

\section{Related Work}

\noindent\textbf{Normal Estimation}.
The classical approaches for normal estimation include Principal Component Analysis (PCA)~\cite{hoppe1992surface} and its refinements~\cite{mitra2003estimating,huang2009consolidation}, which remain popular in many geometric processing tasks.
Other methods~\cite{levin1998approximation,cazals2005estimating,guennebaud2007algebraic}, by introducing new representations of complex surfaces, estimate normals over larger neighborhoods.
However, these techniques often struggle with noisy data and tend to oversmooth geometry when more neighboring points are incorporated.
More recent methods~\cite{guerrero2018pcpnet,ben2019nesti,zhou2020geometry,zhou2022refine,li2022hsurf,li2023NeAF,du2023rethinking,wu2024cmg,xiu2023msecnet} leverage neural networks trained on large, labeled datasets to regress normals for point clouds.
Additionally, other approaches~\cite{lenssen2020deep,cao2021latent,ben2020deepfit,zhu2021adafit,zhou2023improvement,zhang2022geometry,li2022graphfit} focus on predicting pointwise weights through neural networks, with normals subsequently calculated using traditional surface fitting techniques.
However, these methods typically produce unoriented normals and require supervised training with ground truth data for accurate normal predictions.

\noindent\textbf{Normal Orientation}.
For normal orientation, classical methods like Minimum Spanning Tree (MST)~\cite{hoppe1992surface} and its improved variants~\cite{konig2009consistent,schertler2017towards,xu2018towards,jakob2019parallel,metzer2021orienting} rely on propagating orientations through measuring the similarity between neighboring points.
Later, some approaches~\cite{chen2010binary,xiao2023point,xu2023globally} employ volumetric representation techniques to enhance robustness across diverse data, though they often require manual tuning of hyperparameters for different data types.
More recently, researchers have developed deep learning methods~\cite{hashimoto2019normal,wang2022deep,li2023NGLO,li2023shsnet,li2024shsnet-pami} that directly regress oriented normals from point clouds in a data-driven manner.
While these learning-based methods generally outperform traditional data-independent approaches, they often rely heavily on costly labeled training data and struggle with accurately orienting normals in noisy point clouds.
In contrast, our proposed method can learn oriented normals directly from a single noisy point cloud without any labeled data.

\noindent\textbf{Learning Implicit Function from Raw Point Clouds}.
Unlike traditional approaches that train neural networks using supervised signals such as signed distances or occupancy labels, recent works~\cite{atzmon2020sal,ma2020neural,peng2021shape,zhou2022learning,ma2022reconstructing,ma2022surface,chen2023gridpull,zhou2024cap,li2025implicit} have proposed methods to directly learn implicit functions from raw point clouds in an unsupervised manner.
These methods train neural networks to overfit individual point clouds to infer implicit functions without relying on learned priors.
Leveraging gradient constraints~\cite{atzmon2020sal,ma2020neural,chen2023gridpull}, designed priors~\cite{ma2022reconstructing,ma2022surface}, implicit geometric regularization~\cite{gropp2020implicit,giraudot2013noise}, or differentiable Poisson solvers~\cite{peng2021shape}, these techniques can generalize across varying point cloud sizes and accommodate limited input data.
In this work, we build on the neural network's approximation ability and incorporate new techniques for learning signed distance fields.
By applying surface filtering, we aim to mine the underlying geometry of the given data based on implicit field information and accurately infer normals from noisy point clouds.

\begin{figure}[t]
    \centering
    \includegraphics[width=\linewidth]{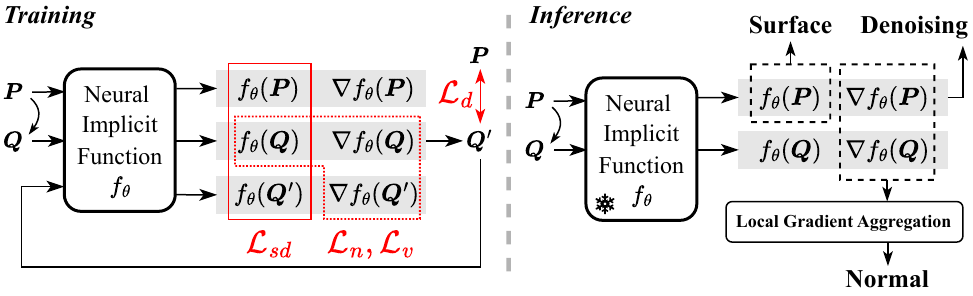}  \vspace{-0.6cm}
    \caption{
        Overview of the proposed method.
        It can be used for different tasks such as surface reconstruction, point cloud denoising, and normal estimation without the need for training labels.
    }
    \label{fig:net}
    \vspace{-0.2cm}
\end{figure}

\section{Method}

\noindent\textbf{Preliminary}.
Implicit representation approaches usually denote surfaces as the level sets of implicit function, \ie,
$\bm{S}_d \!=\! \left \{\bm{x} \in \mathbb{R}^3 ~ \vert ~ f_{\theta}(\bm{x}) \!=\! d \right \}$,
where $f_{\theta} \colon \mathbb{R}^{3} \!\rightarrow\! \mathbb{R}$ is implemented as a neural network with parameter $\theta$.
The implicit function can be learned by overfitting the neural network on individual point clouds.
If the function $f_{\theta}$ is correctly defined by a signed distance field inferred from points, the normal of a point $\bm{p}$ in this implicit field can be obtained by $\bm{n}_{\bm{p}} \!=\! \nabla f_{\theta}(\bm{p})/\norm{\nabla f_{\theta}(\bm{p})}$, where $\norm{\cdot}$ means the Euclidean $L^2$-norm and $\nabla f_{\theta}(\bm{p})$ denotes the gradient at $\bm{p}$.
Specifically, the zero level set $f_{\theta}(\bm{x}) \!=\! 0$ is usually extracted as the object or scene surface $\bm{S}$.
Random points on a level set have specific signed distances, such as $f_{\theta}(\bm{x}) \!<\! 0$ for outside and $f_{\theta}(\bm{x}) \!>\! 0$ for inside.
The gradients on a specific iso-surface should have uniform orientations.
In this work, we aim to apply surface filtering to project noisy points onto the underlying surface defined by the zero level set without supervision of ground truth labels or clean points.
We use the signed distances and normals of noisy points to define the projection path and incorporate rules of the local field.
An overview of our method is provided in Fig.~\ref{fig:net}.

\subsection{Surface Fitting and Filtering}  \label{sec:filter}

We perform surface fitting and point filtering by learning an implicit field from a given noisy point cloud $\bm{P} \!=\! \{ \bm{p}_i|\bm{p}_i \in \mathbb{R}^3\}_{i=1}^N$.
From the perspective of implicit function learning, we aim to construct a signed distance field that minimizes the signed distance of all points to a zero level set, defined as follows:
\begin{equation} \label{eq:sd}
    \arg \min_{f_{\theta}} \frac{1}{N} \sum_{i=1}^{N} |f_{\theta}(\bm{p}_i)|^2 .
\end{equation}
The underlying surface can be fitted by finding the zero level set of the implicit function $f_{\theta}$.
However, directly fitting this surface would force it to pass through all noisy points, resulting in a zero signed distance for each point and thus obtaining a solution to the above equation that fails to accurately represent the desired surface.

From the perspective of data fitting, we typically solve an optimization problem to obtain a best-fitting surface whose distance to all inlier data points is minimized.
The surface $\bm{S}$ to be solved is continuous, and we use its discretization to approximate it.
Let $\bm{\hat{P}} \!=\! \{ \bm{\hat{p}}_i|\bm{\hat{p}}_i \in \mathbb{R}^3\}_{i=1}^{N'},~N' \!>\! N$ denote the discretization of the clean surface, \ie, the point set $\bm{\hat{P}}$ lies on the surface.
In the implicit field space, points $\hat{\bm{p}}_i$ should be located on the zero level set, while noisy points $\bm{p}_i$ may be on a non-zero level set.
By conducting surface fitting using the points in $\bm{P}$, we can solve for the surface points by
\begin{equation} \label{eq:min_d}
    \arg \min_{\hat{\bm{P}}} \frac{1}{N} \sum_{i=1}^{N} \|\hat{\bm{p}}_i - \bm{p}_i\|,
\end{equation}
where each point $\hat{\bm{p}}_i$ is selected for a corresponding point $\bm{p}_i$ based on certain criteria, such as nearest neighbor searching.
However, this distance measure is inadequate because $\hat{\bm{p}}_i$ and $\bm{p}_i$ do not always have a one-to-one correspondence due to the discretization of these points and the interference of noise.
As a result, this approach cannot accurately measure the distance and often yields an over-smoothed geometry or even fails.

\begin{figure}[t]
    \centering
    \includegraphics[width=\linewidth]{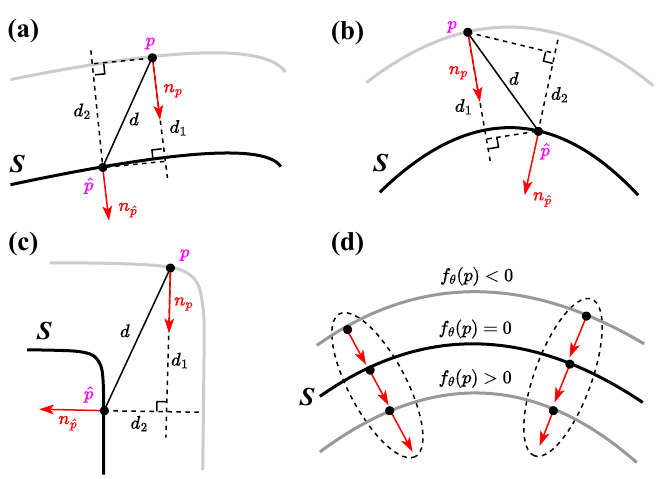}  \vspace{-0.7cm}
    \caption{
        We minimize the distances from noisy points $\bm{p}$ to discrete points $\hat{\bm{p}}$ of the underlying surface for implicit surface fitting and filtering.
        To this end, (a-c) we adopt three distance measures $d$, $d_1$ and $d_2$, and use their sum to handle various cases.
        (d) Meanwhile, we enforce local gradient consistency between adjacent level sets where the noisy points are located.
        Red arrows indicate normals (\ie, gradients).
    }
    \label{fig:dist}
    \vspace{-0.2cm}
\end{figure}

To comprehensively measure the distance error between two points from multiple perspectives, we employ two projection distance measurements using the normals $\bm{n}_{\hat{\bm{p}}_i}$ and $\bm{n}_{\bm{p}_i}$ at points $\hat{\bm{p}}_i$ and $\bm{p}_i$, respectively.
Specifically, these two projection distances are calculated as $d_1 \!=\! | (\hat{\bm{p}}_i - \bm{p}_i) \bm{n}_{\bm{p}_i}^\top|$ and $d_2 \!=\! | (\hat{\bm{p}}_i - \bm{p}_i) \bm{n}_{\hat{\bm{p}}_i}^\top |$, as illustrated in Fig.~\ref{fig:dist}(a-c).
The key insight behind these distance measurements is that if $\hat{\bm{p}}_i$ is the true corresponding surface point of $\bm{p}_i$, then all three distance errors should be minimized.
Taking these projection distances into account, our distance measurement operator for surface fitting from noisy points is defined as
\begin{small}
\begin{equation} \label{eq:dis}
    \mathcal{D}(\hat{\bm{p}}_i, \bm{p}_i) = \frac{1}{N} \sum_{i=1}^{N} \|\hat{\bm{p}}_i - \bm{p}_i\|
                                            + | (\hat{\bm{p}}_i - \bm{p}_i) \bm{n}_{\bm{p}_i}^\top |
                                            + | (\hat{\bm{p}}_i - \bm{p}_i) \bm{n}_{\hat{\bm{p}}_i}^\top |.
\end{equation}
\end{small}
An ideal distance measure is shown in Fig.~\ref{fig:dist}(d), where the level surface is parallel, and the points are correctly matched, resulting in the three distance errors being equal.

\begin{figure}[t]
    \centering
    \includegraphics[width=\linewidth]{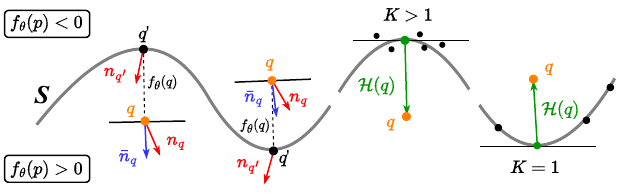}  \vspace{-0.6cm}
    \caption{
        \textit{Left}: computation of $f_{\theta}(\bm{q}) \cdot \bar{\bm{n}}_{\bm{q}}$ and $\bar{\bm{n}}_{\bm{q}} \!=\! (\bm{n}_{\bm{q}} + \bm{n}_{\bm{q}'} ) / || \bm{n}_{\bm{q}} + \bm{n}_{\bm{q}'} ||$.
        Gradients point to the positive side of the signed distance field.
        \textit{Right}: computation of $\mathcal{H}(\bm{q})$ for specific noise and density using different neighborhood scales $K$.
    }
    \label{fig:grad}
    \vspace{-0.1cm}
\end{figure}

\begin{figure}[t]
    \centering
    \includegraphics[width=\linewidth]{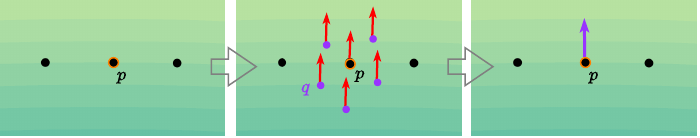}  \vspace{-0.7cm}
    \caption{
        Normal estimation through local gradient aggregation.
    }
    \label{fig:estim}
    \vspace{-0.1cm}
\end{figure}

Next, we introduce the process to obtain the surface points $\hat{\bm{p}}$ and the corresponding point normals $\bm{n}$.
To determine the point $\hat{\bm{p}}$, we first define a new point set $\bm{Q}$, which is generated from the raw point set $\bm{P}$.
This set, $\bm{Q} \!=\! \{ \bm{q}_i \,|\, \bm{q}_i \!\in\! \mathbb{R}^3 \}_{i=1}^N$, is also randomly distributed around the underlying surface.
Since the gradient indicates the direction in which the signed distance from the surface increases most rapidly, moving a point along or against the gradient (depending on the sign of $f_{\theta}$) will allow it to reach its nearest position on the surface.
We thus adopt a point translation operation~\cite{ma2020neural,li2023neuralgf} to project a query point $\bm{q}$ to a new position $\bm{q}'$, where $\bm{q}' \!=\! \bm{q} - f_{\theta}(\bm{q}) \cdot \bm{n}_{\bm{q}}$.
If the implicit function is properly learned, it should provide the correct signed distance $f_{\theta}$ and gradient $\nabla f_{\theta}$ to move the point $\bm{q}$ to its nearest location on the underlying surface.
We then obtain the surface points set $\bm{Q}' \!=\! \{ \bm{q}'_i \,|\, \bm{q}'_i \!=\! \bm{q}_i - f_{\theta}(\bm{q}_i) \cdot \bm{n}_{\bm{q}_i}, ~ \bm{q}_i \in \bm{Q} \}_{i=1}^N$.
Using the raw points in $\bm{P}$ and their nearest points in $\bm{Q}'$, we fit a surface by applying the distance measure operator in Eq.~\eqref{eq:dis}, and the loss function is formulated as
\begin{equation} \label{eq:loss_d}
    \mathcal{L}_d = \frac{1}{N} \sum_{i=1}^{N} \|\bm{q}_i' - \bm{p}_i\|
                    + | (\bm{q}_i' - \bm{p}_i) \bm{n}_{\bm{p}_i}^\top |
                    + | (\bm{q}_i' - \bm{p}_i) \bm{n}_{\bm{q}_i'}^\top |.
\end{equation}
For constructing the noisy point set $\bm{Q}$, we employ a Gaussian based sampling strategy~\cite{atzmon2020sal,cai2020learning}.
Specifically, we first obtain uniformly sampled points $\bm{p}$ from $\bm{P}$, then add Gaussian noise $\mathcal{N}(\bm{p},\sigma^2)$ to each $\bm{p}$, where the standard deviation parameter $\sigma$ is adaptively set based on the distance from $\bm{p}$ to its $\xi$-th nearest neighbor.
In our surface fitting and filtering, we include $\bm{P}$, which together with $\bm{Q}$, to provide more useful information from the raw data.

Based on the surface points $\bm{q}_i' \!\in\! \bm{Q}'$ and their corresponding noisy observations $\bm{q}_i \!\in\! \bm{Q}$ and $\bm{p}_i \!\in\! \bm{P}$, we can define the implicit function learning process using Eq.~\eqref{eq:sd} as follows:
\begin{equation}
    \mathcal{L}_{sd} = \frac{1}{N} \sum_{i=1}^{N} \abs{f_{\theta}(\bm{q}_i')}^2 + \abs{f_{\theta}(\bm{q}_i)}^2 + \abs{f_{\theta}(\bm{p}_i)}^2.
\end{equation}
Since the surface points in $\bm{Q}'$ are located on the zero level set, their signed distances should ideally approach zero.
To enforce this, their signed distances should be assigned a larger weight (\eg, empirically ten times greater) compared to the noisy observations in $\bm{Q}$ and $\bm{P}$, which are normally distributed near the underlying surface and their average signed distances should be zero.

Although $\mathcal{L}_{d}$ and $\mathcal{L}_{sd}$ can guide global surface fitting and filtering, our ablation studies reveal that using these terms alone fails to capture accurate implicit surfaces and point normals in noisy point clouds.
One issue with this approach is that it neglects local geometric details, leading to over-smoothed and noise-sensitive surfaces.
Another significant issue is gradient degradation, which disrupts surface fitting and often accompanies noisy data and complex geometries.
For Eq.~\eqref{eq:loss_d}, we observe that $\nabla f_{\theta} \!=\! 0$ can be an optimal solution, minimizing the function.
This degradation reduces the objective to the original formulation in Eq.~\eqref{eq:min_d}, which implies no valid level set learned by the network, resulting in an inaccurate local distance field and disordered iso-surfaces.
The solution we propose next incorporates local gradient consistency constraints, gradient aggregation between level sets, and local gradient orientation within a level set, effectively addressing these issues.

\subsection{Local Gradient Consistency of Inter-Level}

Inspired by the strategy employed in~\cite{li2023neuralgf}, which constrains directional consistency in a multi-step moving process.
We hope to make the projection $\bm{Q} \!\rightarrow\! \bm{Q}'$ bridge the geometric relationship between noisy points and their corresponding surface points, enhancing the surface filtering accuracy.
We constrain the local gradients of neighboring level sets to have similar directions, as illustrated in Fig.~\ref{fig:dist}(d).
Specifically, we enforce similarity in gradient direction between the initial noisy point $\bm{q}_i$ and its projected point $\bm{q}_i'$.
Recognizing that local gradients between distant level sets may vary, we account for the signed distance in the constraint.
Thus, the confidence-weighted direction distance, used to evaluate gradient consistency between points on neighboring level sets, is formulated as
\begin{equation} \label{eq:loss_con}
  \mathcal{L}_{n} = \frac{1}{N} \sum_{i=1}^{N} \big(1 - \bm{n}_{\bm{q}_i'} ~\bm{n}_{\bm{q}_i}^\top \big) \cdot w_i ~~,
\end{equation}
where $w_i \!=\! {\rm exp}(-\rho \cdot |f_{\theta}(\bm{q}_i)|)$ is an adaptive weight that emphasizes points near the underlying surface based on the predicted distance.
Ablation experiments show that this loss can not only reduce noise impact but also guide the network to generate valid gradients and surface points on level sets.

\begin{figure*}[t]
    \centering
    \includegraphics[width=\linewidth]{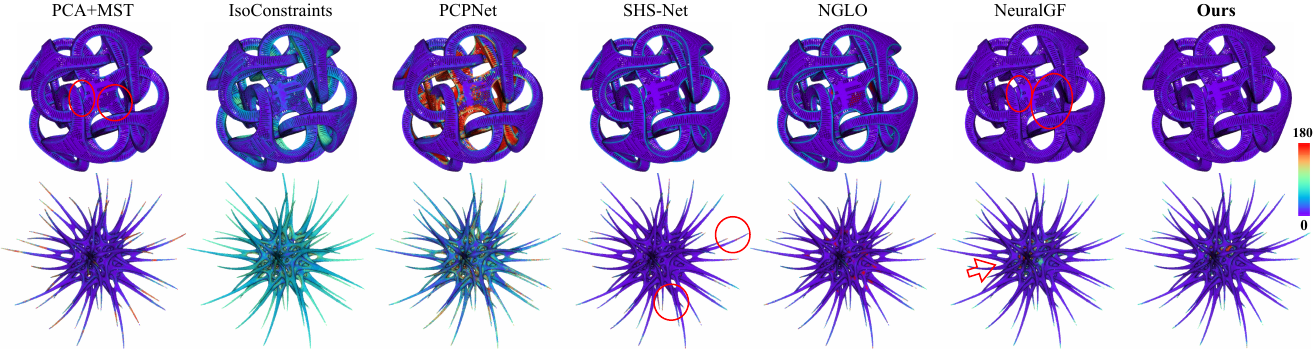}  \vspace{-0.7cm}
    \caption{
    Visual comparison of oriented normals on two point clouds with complex geometry.
    Colors indicate normal errors.
    }
    \label{fig:errorMap_NestPC}
    \vspace{-0.1cm}
\end{figure*}

\begin{figure*}[t]
    \centering
    \includegraphics[width=\linewidth]{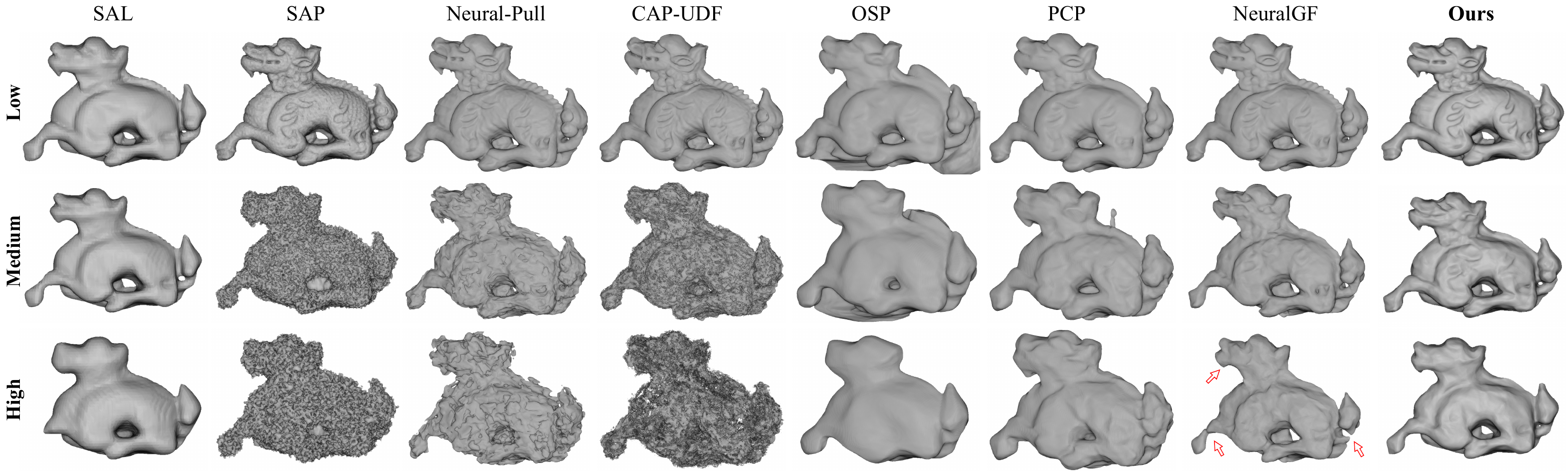}  \vspace{-0.7cm}
    \caption{
    Visual comparison of reconstructed surfaces.
    As the noise increases (from low to high), our method becomes more advantageous.
    }
    \label{fig:surfRecon_netsuke}
    \vspace{-0.1cm}
\end{figure*}

\subsection{Local Gradient Orientation of Intra-Level}

We also focus on the orientation of local gradients at each level set and examine the generation of surface points $\bm{Q}'$ from the raw data.
From the previous equation $\bm{q}' \!=\! \bm{q} - f_{\theta}(\bm{q}) \cdot \bm{n}_{\bm{q}}$, we see that the term $f_{\theta}(\bm{q}) \cdot \bm{n}_{\bm{q}}$ mainly determines the position of the generated surface points.
If the surface point $\bm{q}'$ is known, this term should be as close as possible to $\mathcal{H}(\bm{q}) \!=\! \bm{q} - \bm{q}'$, which we measure by
\begin{equation}
    \mathcal{L}_{v} = \norm{ f_{\theta}(\bm{q}) \cdot \bm{n}_{\bm{q}} - \mathcal{H}(\bm{q}) }.
\end{equation}
Since we only have noisy inputs, we need a robust strategy to approximate $\mathcal{H}(\bm{q})$.
Traditional least squares methods typically use plane fitting within a local neighborhood:
\begin{equation}  \label{eq:plane}
    \mathcal{H}(\bm{q}) = \frac{1}{K} \sum_{k=1}^{K} (\bm{q} - \bm{p}_k), ~~~\bm{p}_k \in \mathbb{K}_{K}(\bm{q}, \bm{P}) ~,
\end{equation}
where $\mathbb{K}_{K}(\bm{q}, \bm{P})$ denotes the set of $K$ nearest points to $\bm{q}$ in $\bm{P}$.
Here, $\mathcal{H}(\bm{q})$ is the oriented vector from the averaged position $\bar{\bm{p}} \!=\! \frac{1}{K} \sum_{k=1}^{K} \bm{p}_k$ to the query point $\bm{q}$.
However, this fixed neighborhood approach is not robust against varying noise levels, density variations and different geometric structures.
In this work, we allow the network model to learn an adaptive neighborhood size to better approximate the surface point by considering multiple scales instead of relying on a fixed neighborhood.
The multi-scale approximation of the surface point $\bm{q}'$ is formulated as
\begin{equation}
    \mathcal{L}_{v} = \sum_{j=1}^{N_K} \norm{ f_{\theta}(\bm{q}) \cdot \bm{n}_{\bm{q}} - \mathcal{H}_j(\bm{q}) } ~,
\end{equation}
where $\mathcal{H}_j(\bm{q})$ is computed using a specific size selected from a scale set $\{K_j\}_{j=1}^{N_K}$, as shown in Fig.~\ref{fig:grad} for specific noise and density.
This formulation reduces the impact of inaccuracies in $\bm{P}$ by utilizing multi-scale local neighbors, thereby inferring the possible correct position of $\bm{q}'$ from multiple corrupted observations of the same local region.

Learning from multiple corrupt observations enhances performance on noisy data, but there remains room for improvement often overlooked by previous works~\cite{zhou2022learning,li2023neuralgf}.
These approaches directly use $\bm{n}_{\bm{q}}$ as the normal of $\bm{q}$, neglecting the inaccuracies in gradients introduced by the local approximation in Eq.~\eqref{eq:plane}.
We take this a step further by rethinking the generation of $\bm{q}'$ from $\bm{q}$ and examining the geometric relationship of their gradients in the implicit field.
For each query point $\bm{q}_i \!\in\! \bm{Q}$, we solve its normal as the sum of the normals at the two endpoints of the projection path, \ie, $\bar{\bm{n}}_{\bm{q}_i} \!=\! (\bm{n}_{\bm{q}_i} + \bm{n}_{\bm{q}_i'}) / || \bm{n}_{\bm{q}_i} + \bm{n}_{\bm{q}_i'} ||$, as illustrated in Fig.~\ref{fig:grad}.
The objective function then becomes the aggregation of errors at each neighborhood size scale:
\begin{equation}
    \mathcal{L}_{v} = \sum_{j=1}^{N_K} \frac{1}{N} \sum_{i=1}^{N}
                        \norm{ f_{\theta}(\bm{q}_i) \cdot \frac{\bm{n}_{\bm{q}_i} + \bm{n}_{\bm{q}_i'}}{|| \bm{n}_{\bm{q}_i} + \bm{n}_{\bm{q}_i'} ||}
                            - \mathcal{H}_j(\bm{q}_i) } .
\end{equation}
Our method learns to identify underlying surface points by using local plane fitting of multi-scale neighbors and replacing $\bm{n}_{\bm{q}}$ with the averaged normal $\bar{\bm{n}}_{\bm{q}}$.
In this way, we can effectively reduce noise-induced errors and avoid potential zero values in $\bm{n}_{\bm{q}}$ due to gradient degradation.

\begin{table*}[t]
\centering
\footnotesize
\setlength{\tabcolsep}{1.1mm}
\caption{
	Oriented normals on PCPNet and FamousShape datasets.
	We achieve better performance even compared with supervised methods.
}
\vspace{-0.3cm}
\label{table:pcpnet_famousShape_o}
\begin{tabular}{l|cccc|cc| >{\columncolor{mygray}} c||  cccc|cc| >{\columncolor{mygray}} c}
	\toprule
	\multirow{3}{*}{Methods} & \multicolumn{7}{c||}{\textbf{PCPNet Dataset}} & \multicolumn{7}{c}{\textbf{FamousShape Dataset}} \\
	\cmidrule(r){2-15}
	& \multicolumn{4}{c|}{Noise} & \multicolumn{2}{c|}{Density} &     & \multicolumn{4}{c|}{Noise} & \multicolumn{2}{c|}{Density} &    \\
	& None & Low & Medium & High  & Stripe & Gradient  & \multirow{-2}{*}{{Average}} & None & Low & Medium & High  & Stripe & Gradient & \multirow{-2}{*}{{Average}} \\
	\midrule
	\multicolumn{15}{l}{\textbf{Supervised}}   \\
	AdaFit~\cite{zhu2021adafit}+MST~\cite{hoppe1992surface}
	& 27.67 & 43.69 & 48.83 & 54.39 & 36.18 & 40.46 &   41.87    & 43.12 & 39.33 & 62.28 & 60.27 & 45.57 & 42.00 &   48.76  \\
	AdaFit~\cite{zhu2021adafit}+SNO~\cite{schertler2017towards}
	& 26.41 & 24.17 & 40.31 & 48.76 & 27.74 & 31.56 &   33.16    & 27.55 & 37.60 & 69.56 & 62.77 & 27.86 & 29.19 &   42.42  \\
	AdaFit~\cite{zhu2021adafit}+ODP~\cite{metzer2021orienting}
	& 26.37 & 24.86 & 35.44 & 51.88 & 26.45 & 20.57 &   30.93    & 41.75 & 39.19 & 44.31 & 72.91 & 45.09 & 42.37 &   47.60  \\
	HSurf-Net~\cite{li2022hsurf}+MST~\cite{hoppe1992surface}
	& 29.82 & 44.49 & 50.47 & 55.47 & 40.54 & 43.15 &   43.99    & 54.02 & 42.67 & 68.37 & 65.91 & 52.52 & 53.96 &   56.24  \\
	HSurf-Net~\cite{li2022hsurf}+SNO~\cite{schertler2017towards}
	& 30.34 & 32.34 & 44.08 & 51.71 & 33.46 & 40.49 &   38.74    & 41.62 & 41.06 & 67.41 & 62.04 & 45.59 & 43.83 &   50.26  \\
	HSurf-Net~\cite{li2022hsurf}+ODP~\cite{metzer2021orienting}
	& 26.91 & 24.85 & 35.87 & 51.75 & 26.91 & 20.16 &   31.07    & 43.77 & 43.74 & 46.91 & 72.70 & 45.09 & 43.98 &   49.37  \\
	PCPNet~\cite{guerrero2018pcpnet}
	& 33.34 & 34.22 & 40.54 & 44.46 & 37.95 & 35.44 &   37.66    & 40.51 & 41.09 & 46.67 & 54.36 & 40.54 & 44.26 &   44.57  \\
	DPGO~\cite{wang2022deep}
	& 23.79 & 25.19 & 35.66 & 43.89 & 28.99 & 29.33 &   31.14    & - & - & - & - & - & - & -  \\
	SHS-Net~\cite{li2023shsnet,li2024shsnet-pami}
	& 10.28 & 13.23 & 25.40 & 35.51 & 16.40 & 17.92 &   19.79    & 21.63 & 25.96 & 41.14 & 52.67 & 26.39 & 28.97 &   32.79  \\
	NGLO~\cite{li2023NGLO}
	& 12.52 & 12.97 & 25.94 & 33.25 & 16.81 & \textbf{9.47}  &   18.49    & \textbf{13.22} & 18.66 & 39.70 & 51.96 & 31.32 & \textbf{11.30} &   27.69  \\

	\midrule
	\multicolumn{15}{l}{\textbf{Unsupervised}}   \\
	PCA~\cite{hoppe1992surface}+MST~\cite{hoppe1992surface}
	& 19.05 & 30.20 & 31.76 & 39.64 & 27.11 & 23.38 &   28.52    & 35.88 & 41.67 & 38.09 & 60.16 & 31.69 & 35.40 &   40.48  \\
	PCA~\cite{hoppe1992surface}+SNO~\cite{schertler2017towards}
	& 18.55 & 21.61 & 30.94 & 39.54 & 23.00 & 25.46 &   26.52    & 32.25 & 39.39 & 41.80 & 61.91 & 36.69 & 35.82 &   41.31  \\
	PCA~\cite{hoppe1992surface}+ODP~\cite{metzer2021orienting}
	& 28.96 & 25.86 & 34.91 & 51.52 & 28.70 & 23.00 &   32.16    & 30.47 & 31.29 & 41.65 & 84.00 & 39.41 & 30.72 &   42.92  \\
	LRR~\cite{zhang2013point}+MST~\cite{hoppe1992surface}
	& 43.48 & 47.58 & 38.58 & 44.08 & 48.45 & 46.77 &   44.82    & 56.24 & 57.38 & 45.73 & 64.63 & 66.35 & 56.65 &   57.83  \\
	LRR~\cite{zhang2013point}+SNO~\cite{schertler2017towards}
	& 44.87 & 43.45 & 33.46 & 45.40 & 46.96 & 37.73 &   41.98    & 59.78 & 60.18 & 45.02 & 71.37 & 62.78 & 59.90 &   59.84  \\
	LRR~\cite{zhang2013point}+ODP~\cite{metzer2021orienting}
	& 28.65 & 25.83 & 36.11 & 53.89 & 26.41 & 23.72 &   32.44    & 39.97 & 42.17 & 48.29 & 88.68 & 44.92 & 47.56 &   51.93  \\
	IsoConstraints~\cite{xiao2023point}
	& 24.42 & 26.52 & 87.30 & 94.99 & 28.69 & 32.02 &   48.99    & 38.23 & 41.59 & 83.11 & 93.07 & 42.47 & 49.68 &   58.03  \\
	NeuralGF~\cite{li2023neuralgf}
	& 10.60 & 18.30 & 24.76 & 33.45 & 12.27 & 12.85 &   18.70    & 16.57 & 19.28 & 36.22 & 50.27 & 17.23 & 17.38 &   26.16  \\
	Ours
	& \textbf{9.71} & \textbf{11.99} & \textbf{24.39} & \textbf{32.74} & \textbf{11.30} & {11.84} &   \textbf{17.00}
	&        {13.71} & \textbf{18.40} & \textbf{34.97} & \textbf{49.25} & \textbf{14.35} & {13.76} &   \textbf{24.07}  \\
	\bottomrule
\end{tabular} 
\end{table*}
\begin{table}[t]
    \footnotesize
    \centering
    \setlength{\tabcolsep}{0.9mm}
    \caption{
      Comparison of oriented normals using PGP-$70^\circ$ (\%) under medium noise on PCPNet and FamousShape datasets.
    }
    \vspace{-0.3cm}
    \resizebox{\linewidth}{!}{
    \begin{tabular}{l|ccccccccc}
      \toprule
      Dataset       & \tabincell{c}{HSurf-Net\\+ODP}  & \tabincell{c}{Iso-\\Cons.}  & PCPNet    & NGLO      & SHS-Net  & NeuralGF  & Ours            \\
      \midrule
      PCPNet        & 90.28                           & 58.54                       & 89.47     & 94.59     & 94.59    & 94.67     & \textbf{94.96}  \\
      Famous.       & 87.57                           & 57.83                       & 87.19     & 91.10     & 90.90    & 92.87     & \textbf{93.14}  \\
      \bottomrule
    \end{tabular} }
    \label{tab:PGP_o}
\end{table}
\begin{table}[t]
    \footnotesize
    \centering
    \setlength{\tabcolsep}{0.8mm}
    \caption{
        Oriented normals on sparse point cloud data. 
        IsoConstraints~\cite{xiao2023point} and GCNO~\cite{xu2023globally} are traditional methods.
    }
    \vspace{-0.3cm}
    \resizebox{\linewidth}{!}{
    \begin{tabular}{c|cccccccc}
    \toprule
                  & \tabincell{c}{HSurf-Net\\+ODP}  & \tabincell{c}{Iso-\\Cons.}        & GCNO   & PCPNet  & NGLO   & SHS-Net  & NeuralGF  & Ours             \\  
    \midrule
    3K          & 63.88                           & 40.01                             & 33.40  & 53.13   & 32.65  & 37.31    & 25.54     & \textbf{24.03}   \\  
    5K          & 62.51                           & 37.45                             & 41.24  & 48.48   & 28.34  & 32.64    & 24.35     & \textbf{21.80}   \\  
    \bottomrule
    \end{tabular} }
    \label{tab:sparse_o}
\end{table}
\begin{table}[t]
    \footnotesize
    \centering
    \setlength{\tabcolsep}{1.3mm}
    \caption{
        Oriented normals on SceneNN and ScanNet datasets.
    }
    \vspace{-0.3cm}
    \resizebox{\linewidth}{!}{
    \begin{tabular}{l|cccccccc}
    \toprule
    \makebox[0.13\linewidth][c]{Dataset}      & \tabincell{c}{HSurf-Net\\+ODP}  & PCPNet  & NGLO   & SHS-Net  & NeuralGF  & Ours             \\   
    \midrule
    \multicolumn{7}{l}{\textbf{SceneNN}}   \\
    Clean                                     & 51.85                           & 70.70   & 48.52  & 78.71    & 47.80     & \textbf{44.82}   \\   
    Noisy                                     & 50.24                           & 70.82   & 45.42  & 77.60    & 48.69     & \textbf{40.66}   \\   
    Average                                   & 51.05                           & 70.76   & 46.97  & 78.16    & 48.24     & \textbf{42.74}   \\   
    \midrule
    \midrule
    \textbf{ScanNet}                          & 49.34                           & 68.10   & 39.40  & 74.36    & 39.10     & \textbf{37.09}   \\   
    \bottomrule
    \end{tabular} }
    \label{tab:scene}
\end{table}

\noindent\textbf{Final Loss}.
Our final loss function of learning normals from noisy points is defined as
\begin{equation} \label{eq:loss}
  \mathcal{L} = \mathcal{L}_{sd} + \mathcal{L}_{v} + \lambda_1 \mathcal{L}_{d} + \lambda_2 \mathcal{L}_{n} ~~,
\end{equation}
where the weight factors $\lambda_1$ and $\lambda_2$ are first set empirically and then fine-tuned based on experimental results.

\subsection{Local Gradient Aggregation of Inter-Level}  \label{sec:normal_infer}

We train the network model to overfit on a given single point cloud $\bm{P}$.
To infer the normal of a point $\bm{p} \!\in\! \bm{P}$ using the learned model, we first use the combination of $\bm{P}$ and $\bm{Q}$ as the input and derive the gradients of all points.
The additional set $\bm{Q}$ is used to fully explore the possible true spatial positions of noisy points $\bm{P}$ corresponding to the underlying surface.
We then search for the $\kappa$ nearest neighbors of $\bm{p}$ in the point set $\{\bm{P}, \bm{Q}\}$, as shown in Fig.~\ref{fig:estim}, and the final normal of $\bm{p}$ is calculated as the weighted average of the normal at $\bm{p}$ and its neighboring point normals $\bar{\bm{n}}_i'$, as follows:
\begin{equation}  \label{eq:normal}
    \bm{n}_{\bm{p}} = \frac{1}{\kappa+1} \left( \bar{\bm{n}}_{\bm{p}} + \sum_{i=1}^{\kappa} \bar{\bm{n}}_i' \cdot \mu_i \right),
\end{equation}
where
$\mu_i \!=\! {\rm exp}\left(-\delta_i - \phi(\bar{\bm{n}}_i', \bar{\bm{n}}_{\bm{p}}) \right)$,
and $\delta_i$ is the Euclidean distance between point $\bm{p}$ and its neighboring points.
$\phi(\bar{\bm{n}}_i', \bar{\bm{n}}_{\bm{p}}) \!=\! \left( (1 - \bar{\bm{n}}_i' ~\bar{\bm{n}}_{\bm{p}}^\top)/(1 - {\rm cos}~\vartheta) \right)^2$,
where $\vartheta$ is a given angle.
Given the nearest neighbors, the term $\mu$ adaptively assigns higher weight to neighboring points that are closer to point $\bm{p}$ or have a small normal angle with it.
Ablation experiments show that this strategy is effective in improving the robustness of the algorithm in various cases.

\section{Experiments}

\noindent\textbf{Implementation}.
We employ a simple neural network similar to that used in~\cite{atzmon2020sal,ma2020neural,li2023neuralgf}.
It consists of eight linear layers and includes a skip connection.
We also use the geometric network initialization from~\cite{atzmon2020sal}.
In all evaluation experiments, the network structure and loss function components remain consistent.
The neighborhood scale set $\{K_j\}_{j=1}^{N_K}$ is chosen as $\{1, \mathcal{K}/2, \mathcal{K}\}$, with $N_K \!=\! 3$, $\mathcal{K} \!=\! 8$, and we set the parameters $\lambda_2 \!=\! 0.01$, $\rho \!=\! 60$, $\kappa \!=\! 8$, and $\vartheta \!=\! \pi/12$.
The hyperparameters $\xi$ and $\lambda_1$ are adjusted based on specific datasets.
The number of points $N$ used in each training iteration is set to $5000$.
These hyperparameters are fixed for all test samples within each dataset, and no per-shape tuning is needed.
As in~\cite{guerrero2018pcpnet,li2022hsurf,li2023shsnet}, we evaluate the normal estimation results using Root Mean Squared Error (RMSE) and Percentage of Good Points (PGP) with various thresholds.
Our evaluation covers multiple noise levels (Gaussian standard deviations of 0.12\% (low), 0.6\% (medium), and 1.2\% (high) of the diagonal of object's bounding box), `gradient' representing density variations, and `stripe' simulating occlusions.
\textit{More experimental results are provided in the supplementary material.}

\begin{table*}[t]
\centering
\footnotesize
\setlength{\tabcolsep}{1.4mm}
\caption{
	Unoriented normals on PCPNet and FamousShape datasets.
	We achieve better performance compared with unsupervised methods.
}
\vspace{-0.3cm}
\label{table:pcpnet_famousShape}
\begin{tabular}{l|cccc|cc| >{\columncolor{mygray}} c||  cccc|cc| >{\columncolor{mygray}} c}
	\toprule
	\multirow{3}{*}{Methods} & \multicolumn{7}{c||}{\textbf{PCPNet Dataset}} & \multicolumn{7}{c}{\textbf{FamousShape Dataset}} \\
	\cmidrule(r){2-15}
	& \multicolumn{4}{c|}{Noise} & \multicolumn{2}{c|}{Density} &     & \multicolumn{4}{c|}{Noise} & \multicolumn{2}{c|}{Density} &    \\
	& None & Low & Medium & High  & Stripe & Gradient  & \multirow{-2}{*}{{Average}} & None & Low & Medium & High  & Stripe & Gradient & \multirow{-2}{*}{{Average}} \\
	\midrule
	\multicolumn{15}{l}{\textbf{Supervised}}   \\
	DeepFit~\cite{ben2020deepfit}                & 6.51  & 9.21  & 16.73 & 23.12 & 7.92  & 7.31  &  11.80    & 11.21 & 16.39 & 29.84 & 39.95 & 11.84 & 10.54 &   19.96  \\
	Zhang \etal~\cite{zhang2022geometry}         & 5.65  & 9.19  & 16.78 & 22.93 & 6.68  & 6.29  &  11.25    & 9.83 & 16.13 & 29.81 & 39.81 & 9.72 & 9.19 &   19.08  \\
	AdaFit~\cite{zhu2021adafit}                  & 5.19  & 9.05  & 16.45 & 21.94 & 6.01  & 5.90  &  10.76    & 9.09 & 15.78 & 29.78 & 38.74 & 8.52 & 8.57 &   18.41  \\
	GraphFit~\cite{li2022graphfit}               & 5.21  & 8.96  & 16.12 & 21.71 & 6.30  & 5.86  &  10.69    & 8.91 & 15.73 & 29.37 & 38.67 & 9.10 & 8.62 &   18.40  \\
	NeAF~\cite{li2023NeAF}                       & 4.20  & 9.25  & 16.35 & 21.74 & 4.89  & 4.88  &  10.22    & 7.67 & 15.67 & 29.75 & 38.76 & 7.22 & 7.47 &   17.76  \\
	HSurf-Net~\cite{li2022hsurf}                 & 4.17  & 8.78  & 16.25 & 21.61 & 4.98  & 4.86  &  10.11    & 7.59 & 15.64 & 29.43 & 38.54 & 7.63 & 7.40 &   17.70  \\
	NGLO~\cite{li2023NGLO}                       & 4.06  & 8.70  & 16.12 & 21.65 & 4.80  & 4.56  &  9.98     & 7.25 & 15.60 & 29.35 & 38.74 & 7.60 & 7.20 &   17.62  \\
	SHS-Net~\cite{li2023shsnet,li2024shsnet-pami} & 3.95  & 8.55  & 16.13 & 21.53 & 4.91  & 4.67  &  9.96     & 7.41 & 15.34 & 29.33 & 38.56 & 7.74 & 7.28 &   17.61  \\
	Du \etal~\cite{du2023rethinking}             & 3.85  & 8.67  & 16.11 & 21.75 & 4.78  & 4.63  &  9.96     & 6.92 & 15.05 & 29.49 & 38.73 & 7.19 & 6.92 &   17.38  \\
	CMG-Net~\cite{wu2024cmg}                     & 3.87  & 8.45  & 16.08 & 21.89 & 4.85  & 4.45  &  9.93     & 7.07 & 14.83 & 29.04 & 38.93 & 7.43 & 7.03 &   17.39  \\
	MSECNet~\cite{xiu2023msecnet}                & 3.84  & 8.74  & 16.10 & 21.05 & 4.34  & 4.51  &  9.76     & 6.85 & 15.60 & 29.22 & 38.13 & 6.64 & 6.65 &   17.18  \\
	\midrule
	\multicolumn{15}{l}{\textbf{Unsupervised}}   \\
	CAP-UDF~\cite{zhou2024cap}                   & \textbf{7.59}  & 11.99 & 37.69 & 47.64 & \textbf{8.26}  & \textbf{7.36}  &  20.09    & 14.34 & 21.62 & 50.43 & 55.33 & 13.31 & 13.45 &   28.08  \\
	Boulch \etal~\cite{boulch2012fast} 	         & 11.80 & 11.68 & 22.42 & 35.15 & 13.71 & 12.38 &  17.86    & 19.00 & 19.60 & 36.71 & 50.41 & 20.20 & 17.84 &   27.29  \\
	PCV~\cite{zhang2018multi} 	                 & 12.50 & 13.99 & 18.90 & 28.51 & 13.08 & 13.59 &  16.76    & 21.82 & 22.20 & 31.61 & 46.13 & 20.49 & 19.88 &   27.02  \\
	Jet~\cite{cazals2005estimating}              & 12.35 & 12.84 & 18.33 & 27.68 & 13.39 & 13.13 &  16.29    & 20.11 & 20.57 & 31.34 & 45.19 & 18.82 & 18.69 &   25.79  \\
	PCA~\cite{hoppe1992surface} 	             & 12.29 & 12.87 & 18.38 & 27.52 & 13.66 & 12.81 &  16.25    & 19.90 & 20.60 & 31.33 & 45.00 & 19.84 & 18.54 &   25.87  \\
	LRR~\cite{zhang2013point} 	                 & 9.63  & 11.31 & 20.53 & 32.53 & 10.42 & 10.02 &  15.74    & 17.68 & 19.32 & 33.89 & 49.84 & 16.73 & 16.33 &   25.63  \\
	NeuralGF~\cite{li2023neuralgf}               & 7.89  & 9.85  & 18.62 & 24.89 & 9.21  & 9.29  &  13.29    & 13.74 & 16.51 & 31.05 & 40.68 & 13.95 & 13.17 &   21.52  \\
	Ours
	&        {7.60}  & \textbf{9.45}  & \textbf{16.87} & \textbf{22.49} &        {8.52}  &        {8.55}  &   \textbf{12.25}
	& \textbf{11.90} & \textbf{15.84} & \textbf{29.90} & \textbf{39.08} & \textbf{11.82} & \textbf{11.36} &   \textbf{19.98}   \\
	\bottomrule
\end{tabular} 
\end{table*}
\begin{table}[t]
    \footnotesize
    \centering
    \setlength{\tabcolsep}{0.5mm}
    \caption{
      Comparison of unoriented normals using PGP-$20^\circ$ (\%) under high noise on PCPNet and FamousShape datasets.
    }
    \vspace{-0.3cm}
    \resizebox{\linewidth}{!}{
    \begin{tabular}{l|cccccccccc}
      \toprule
      Dataset       & Jet    & PCA    & PCV    & LRR    & Boulch \etal  & CAP-UDF  & NeuralGF  &  Ours             \\
      \midrule
      PCPNet        & 64.60  & 65.06  & 62.79  & 52.33  & 44.03         & 24.84    & 70.02     & \textbf{74.65}    \\
      Famous.       & 27.46  & 28.05  & 25.55  & 19.29  & 17.52         & 12.23    & 37.32     & \textbf{42.62}    \\
      \bottomrule
    \end{tabular} }
    \label{tab:PGP_uo}
\end{table}

\begin{table}[t]
    \footnotesize
    \centering
    \setlength{\tabcolsep}{.8mm}
    \caption{
        Surface reconstruction on the SRB dataset.
    }
    \vspace{-0.3cm}
    \resizebox{\linewidth}{!}{
    \begin{tabular}{c|cccccc}
        \toprule
                           & SAP                      & Neural-Pull                     & NeuralGF                       & CAP-UDF                    & IF                       & Ours             \\
        \midrule
        ${\rm CD}_{L1}$    & 0.4787                   & 0.2845                          & 0.2623                         & 0.2766                     & 0.2519                   & \textbf{0.2518}  \\
        F-Score            & 0.9383                   & 0.9689                          & 0.9758                         & 0.9760                     & 0.9782                   & \textbf{0.9786}  \\
        \bottomrule
    \end{tabular} }
    \label{tab:srb}
    \vspace{-0.2cm}
\end{table}
\begin{table}[t]
    \centering
    \footnotesize
    \setlength{\tabcolsep}{1.2mm}
    \caption{
        Surface reconstruction on the 3D Scene dataset.
    }
    \vspace{-0.3cm}
    \resizebox{\linewidth}{!}{
    \begin{tabular}{l|ccc|ccc}
        \toprule
                                          & \multicolumn{3}{c|}{Stonewall}                        & \multicolumn{3}{c}{Lounge}                               \\ 
        \hline
        Metric                            & ${\rm CD}_{L2}$ & ${\rm CD}_{L1}$ & NC                & ${\rm CD}_{L2}$    & ${\rm CD}_{L1}$ & NC                \\ 
        \midrule
        Neural-Pull~\cite{ma2020neural}   & 27.2995         & 3.0477          & 0.8222            & 0.3172             & 0.2350          & 0.8949            \\ 
        OSP~\cite{ma2022reconstructing}   & 0.7241          & 0.5226          & 0.8878            & 7.3628             & 1.6020          & 0.6828            \\
        SAP~\cite{peng2021shape}          & 0.5499          & 0.2988          & 0.8599            & 0.1372             & 0.2221          & 0.8480            \\ 
        NeuralGF~\cite{li2023neuralgf}    & 0.0534          & 0.0934          & 0.9469            & 0.1428             & 0.1658          & 0.9059            \\
        CAP-UDF~\cite{zhou2024cap}        & 0.0107          & 0.0795          & 0.9403            & \textbf{0.0221}    & \textbf{0.1086} & 0.8903            \\ 
        IF~\cite{li2025implicit}          & 0.1222          & 0.1998          & 0.9238            & 0.1046             & 0.1519          & 0.8979            \\ 
        Ours                              & \textbf{0.0093} & \textbf{0.0777} & \textbf{0.9527}   & 0.1158             & 0.1531          & \textbf{0.9093}   \\ 
        \bottomrule
    \end{tabular} }
    \label{tab:3dscene}
\end{table}

\begin{table*}[t]
\footnotesize
\centering
\setlength{\tabcolsep}{2.5mm}
\caption{
    Quantitative results of point cloud denoising on the PointCleanNet dataset.
    Our method is unsupervised, while all baselines are supervised.
    Bold and underlined denote the best and second best, respectively.
}
\vspace{-0.3cm}
\begin{tabular}{c|l|cccccc|cccccc}
    \toprule
    \multicolumn{2}{c|}{Resolution} & \multicolumn{6}{c|}{10K (Sparse)} & \multicolumn{6}{c}{50K (Dense)}             \\
    \midrule
    \multicolumn{2}{c|}{Noise}      & \multicolumn{2}{c}{1\%} & \multicolumn{2}{c}{2\%} & \multicolumn{2}{c|}{3\%}
                & \multicolumn{2}{c}{1\%} & \multicolumn{2}{c}{2\%} & \multicolumn{2}{c}{3\%}     \\
    \midrule
    \multicolumn{2}{c|}{Metric ($\times10^4$)}     & CD & P2M & CD & P2M & CD & P2M            & CD & P2M & CD & P2M & CD & P2M      \\
    \midrule

    \parbox[t]{2.5mm}{\multirow{7}{*}{\rotatebox[origin=c]{90}{\textbf{Supervised}}}}

    & PointCleanNet~\cite{rakotosaona2020pointcleannet}     & 3.853 & 1.711 & 8.796 & 4.036 & 14.531 & 7.505  & 1.275 & 0.431 & 1.875 & 0.762 & 3.204 & 1.650     \\
    & DMRDenoise~\cite{luo2020differentiable}       & 6.573 & 2.867 & 7.136 & 3.066 & 8.141 & 3.496   & 1.567 & 0.548 & 2.016 & 0.734 & 2.973 & 1.196     \\
    & Pointfilter~\cite{zhang2020pointfilter}       & 3.291 & 1.434 & 5.887 & 2.756 & 7.814 & 4.085   & 1.046 & 0.302 & 1.661 & 0.732 & 2.532 & 1.256     \\
    & ScoreDenoise~\cite{luo2021score}              & 3.369 & 1.264 & 5.132 & 1.798 & 6.776 & 2.864   & 1.066 & 0.312 & 1.659 & 0.567 & 2.494 & 1.057     \\
    & PD-Flow~\cite{mao2022pd}                      & 3.245 & \textbf{0.941} & 4.626 & \textbf{1.410} & \textbf{5.956} & \textbf{2.150}   & \underline{0.968} & 0.252 & 1.660 & 0.691 & 2.470 & \underline{0.922}     \\
    & DeepPSR~\cite{chen2022deep}                   & 3.141 & 1.434 & 4.929 & 1.998 & 6.140 & 2.594   & 0.990 & 0.280 & 1.548 & 0.620 & \textbf{2.123} & 0.981     \\
    & IterativePFN~\cite{de2023iterativepfn}        & \textbf{2.621} & 1.061 & \textbf{4.437} & \underline{1.544} & 6.021 & 2.284   & \textbf{0.913} & \underline{0.243} & \textbf{1.252} & \textbf{0.419} & 2.546 & 1.040     \\

    \midrule
    \multicolumn{2}{c|}{Ours}                       & \underline{2.660} & \underline{1.008} & \underline{4.622} & 1.553 & \underline{5.984} & \underline{2.215}   & 0.991 & \textbf{0.241} & \underline{1.517} & \underline{0.461} & \underline{2.168} & \textbf{0.793}     \\

    \bottomrule
\end{tabular} 
\label{tab:denoising}
\end{table*}

\begin{table*}[t]
\centering
\footnotesize
\setlength{\tabcolsep}{1mm}
\caption{
    Ablations for unoriented and oriented normal estimation on the FamousShape dataset.
    We decompose $\mathcal{L}_d$ into $\mathcal{L}_{ld}$ and $\mathcal{L}_{pd}$.
}
\vspace{-0.3cm}
\label{tab:ablation}
\begin{tabular}{ll|cccc|cc| >{\columncolor{mygray}} c||  cccc|cc| >{\columncolor{mygray}} c}
    \toprule
    & \multirow{3}{*}{Category} & \multicolumn{7}{c||}{\textbf{Unoriented Normal RMSE}} & \multicolumn{7}{c}{\textbf{Oriented Normal RMSE}} \\
    \cmidrule(r){3-16}
    &
    & \multicolumn{4}{c|}{Noise} & \multicolumn{2}{c|}{Density} &    & \multicolumn{4}{c|}{Noise} & \multicolumn{2}{c|}{Density} &  \\
    &
    & None & Low & Medium & High & Stripe & Gradient & \multirow{-2}{*}{{Average}}
    & None & Low & Medium & High & Stripe & Gradient & \multirow{-2}{*}{{Average}} \\
    \midrule
    \multirow{10}{*}{\textbf{(a)}}
    & $\mathcal{L}_{ld}$
    & 25.19 & 33.08 & 38.42 & 46.76 & 28.80 & 25.51 &   32.96      & 37.51 & 67.44 & 84.01 & 69.70 & 45.80 & 35.95 &   56.73   \\
    & $\mathcal{L}_{ld}+\mathcal{L}_{pd}+\mathcal{L}_{sd}$
    & 15.74 & 23.08 & 54.64 & 56.10 & 14.33 & 15.53 &   29.90      & 22.11 & 39.46 & 88.36 & 92.84 & 17.97 & 21.94 &   47.11   \\
    & $\mathcal{L}_{ld}+\mathcal{L}_{pd}+\mathcal{L}_{sd}+\mathcal{L}_{v}$
    & 16.98 & 16.60 & 33.83 & 41.23 & 18.85 & 18.37 &   24.31      & 26.56 & 23.91 & 55.90 & 83.79 & 28.98 & 27.41 &   41.09   \\
    & $\mathcal{L}_{ld}+\mathcal{L}_{pd}+\mathcal{L}_{sd}+\mathcal{L}_{n}$
    & 13.41 & 16.70 & 31.11 & 40.48 & 13.00 & 12.78 &   21.25      & 19.42 & 19.48 & 36.03 & 50.38 & 14.76 & \textbf{13.29} &   25.56   \\
    & $\mathcal{L}_{ld}+\mathcal{L}_{pd}+\mathcal{L}_{n}+\mathcal{L}_{v}$
    & 12.00 & 15.87 & 30.00 & 39.30 & 11.95 & 13.14 &   20.38      & 14.66 & 18.89 & 34.96 & 55.00 & 15.12 & 16.45 &   25.85   \\
    & $\mathcal{L}_{ld}+\mathcal{L}_{sd}+\mathcal{L}_{n}+\mathcal{L}_{v}$
    & 12.61 & 15.59 & 30.19 & 39.13 & 12.38 & 11.69 &   20.27      & 16.06 & 18.30 & 38.79 & 48.26 & 15.70 & 14.99 &   25.35   \\
    & $\mathcal{L}_{ld}+\mathcal{L}_{n}+\mathcal{L}_{v}$
    & 12.28 & 15.63 & 29.88 & 39.26 & 12.18 & 12.19 &   20.24      & 14.98 & 18.25 & \textbf{34.62} & 50.82 & 14.95 & 21.90 &   25.92   \\
    & $\mathcal{L}_{pd}+\mathcal{L}_{n}+\mathcal{L}_{v}$
    & 11.88 & 15.72 & 29.97 & \textbf{39.05} & 12.24 & 12.14 &   20.17      & 13.32 & 18.34 & 34.87 & 48.44 & 15.08 & 23.17 &   25.54   \\
    & $\mathcal{L}_{sd}+\mathcal{L}_{n}+\mathcal{L}_{v}$
    & 12.49 & \textbf{15.47} & 29.94 & 39.19 & 12.49 & 12.11 &   20.28      & 15.06 & \textbf{17.15} & 35.25 & \textbf{48.27} & 15.44 & 16.62 &   24.63   \\
    & $\mathcal{L}_{pd}+\mathcal{L}_{sd}+\mathcal{L}_{n}+\mathcal{L}_{v}$
    & \textbf{11.81} & 15.73 & 29.97 & 39.08 & 12.01 & 11.43 &   20.00      & \textbf{13.31} & 18.36 & 34.89 & 48.87 & 15.15 & 14.07 &   24.11   \\
    \midrule
    \multirow{1}{*}{\textbf{(b)}}
    & w/o Aggregation
    & 12.05 & 16.02 & 30.04 & 39.18 & 11.96 & 11.51 &   20.13      & 13.82 & 18.53 & 35.11 & 49.34 & 14.46 & 13.85 &   24.19   \\
    \midrule
    \multirow{2}{*}{\textbf{(c)}}
    & $\mathcal{K}=4$
    & 11.89 & 16.00 & 30.22 & 39.51 & 17.45 & \textbf{11.17} &   21.04      & 14.80 & 19.07 & 35.16 & 51.31 & 21.49 & 13.65 &   25.91   \\
    & $\mathcal{K}=16$
    & 13.71 & 15.77 & \textbf{29.86} & 39.32 & 12.11 & 13.49 &   20.71      & 20.41 & 18.44 & 34.90 & 51.19 & \textbf{14.14} & 17.87 &   26.16   \\
    \midrule
    & \multirow{1}{*}{\textbf{Full}}
    & 11.90 & 15.84 & 29.90 & 39.08 & \textbf{11.82} & 11.36 &   \textbf{19.98}      & 13.71 & 18.40 & 34.97 & 49.25 & 14.35 & 13.76 &   \textbf{24.07}   \\
    \bottomrule
\end{tabular} 
\end{table*}

\subsection{Normal Estimation}

\noindent\textbf{Comparison of Oriented Normal.}
Our approach requires no training labels and learns solely from raw data.
We compare our method with both supervised and unsupervised methods, including end-to-end and two-stage pipeline approaches.
In Table~\ref{table:pcpnet_famousShape_o}, we report quantitative evaluation results on the PCPNet~\cite{guerrero2018pcpnet} and FamousShape~\cite{li2023shsnet} datasets.
Our method achieves superior performance across most data categories (in terms of different noise levels and density variations) and delivers the best average results, even compared to supervised approaches.
Quantitative comparisons of PGP at a threshold of $70^\circ$ are reported in Table~\ref{tab:PGP_o}, indicating that our method provides more accurate normals for a higher proportion of points.

In Table~\ref{tab:sparse_o}, we present evaluation results on sparse point cloud data.
These point sets are sparse versions of the FamousShape dataset~\cite{li2023shsnet}, each containing only $3000$ and $5000$ points.
The quantitative comparison results demonstrate that our method achieves the lowest error on these sparse point sets.
We further evaluate our approach on real-world scanned datasets to assess its generalization capability.
Table~\ref{tab:scene} provides quantitative results on the SceneNN~\cite{hua2016scenenn} and ScanNet~\cite{dai2017scannet} datasets, where our method outperforms baselines, demonstrating a stronger ability to handle real-world data.

\noindent\textbf{Comparison of Unoriented Normal.}
In this evaluation, we use our oriented normals to compare with baselines but ignore normal orientations, which are often challenging to determine.
In Table~\ref{table:pcpnet_famousShape}, we report quantitative results on the PCPNet and FamousShape datasets.
Most existing methods for unoriented normal estimation rely on supervised training with ground truth normals.
We evaluate both supervised and unsupervised methods, and our approach outperforms in most data categories (varying noise and density) across both datasets, achieving the highest average results among unsupervised methods.
Notably, CAP-UDF~\cite{zhou2022learning,zhou2024cap} performs well on noise-free point clouds but struggles with noisy data.
In contrast, our method demonstrates a significant advantage on noisy data.
The quantitative comparisons of PGP at a threshold of $20^\circ$ are reported in Table~\ref{tab:PGP_uo}, showing that our method delivers more accurate normals for a larger proportion of points.

\subsection{Surface Reconstruction}

The zero level set of our learned implicit function can be extracted as the object surface using the marching cubes algorithm~\cite{lorensen1987marching}.
We compare our surface reconstruction performance with other implicit representation methods, including SAL~\cite{atzmon2020sal}, SAP~\cite{peng2021shape}, Neural-Pull~\cite{ma2020neural}, CAP-UDF~\cite{zhou2024cap}, OSP~\cite{ma2022reconstructing}, PCP~\cite{ma2022surface}, IF~\cite{li2025implicit}, and NeuralGF~\cite{li2023neuralgf}.
As shown in Fig.~\ref{fig:surfRecon_netsuke}, we present a visual comparison of reconstructed surfaces from point clouds at varying noise levels.
Our method shows a clear advantage in handling noisy data, producing cleaner and more complete structures than baseline methods.
For surface reconstruction from real-world data, we follow previous works~\cite{zhou2024cap,ma2022surface,li2025implicit} and evaluate on the SRB dataset~\cite{williams2019deep} and 3D Scene dataset~\cite{zhou2013dense}, using Chamfer distance (CD), Normal Consistency (NC), and F-Score metrics.
The quantitative results reported in Table~\ref{tab:srb} and Table~\ref{tab:3dscene} show that our method achieves the highest accuracy on the SRB dataset and in most cases of the 3D Scene dataset.

\subsection{Point Cloud Denoising}

In Sec.~\ref{sec:filter}, we use the raw point cloud $\bm{P}$ to construct a new sample set $\bm{Q}$, and obtain the corresponding surface point set $\bm{Q}'$ through filtering.
For point cloud denoising, we take all points in the raw data $\bm{P}$ as input to the trained model.
By applying the transformation $\bm{P}' \!=\! \{ \bm{p}'_i \mid \bm{p}'_i \!=\! \bm{p}_i - f_{\theta}(\bm{p}_i) \cdot \bm{n}_{\bm{p}_i}, ~\bm{p}_i \in \bm{P} \}_{i=1}^N$, the new generated points $\bm{P}'$ should ideally lie on the underlying clean surface.
Following prior works~\cite{luo2021score,chen2022deep}, we evaluate our denoising performance on the PointCleanNet dataset~\cite{rakotosaona2020pointcleannet}, a standard benchmark that includes two resolution levels (10K and 50K points) and three noise levels (scales of 1\%, 2\%, and 3\% of the shape bounding sphere's radius).
We also use Chamfer distance (CD) and point-to-mesh distance (P2M) as metrics to evaluate the denoised point clouds.
The quantitative comparison results are reported in Table~\ref{tab:denoising}.
Unlike existing learning-based denoising methods, which typically require clean surface data for supervised training, our unsupervised method achieves comparable performance to these supervised baselines.
Additionally, our network is lightweight, with approximately 461K parameters ($14\%$ of IterativePFN's 3.2M parameters~\cite{de2023iterativepfn}).
This experiment demonstrates that our method can effectively recover underlying surfaces from noisy point clouds.

\subsection{Ablation Studies}  \label{sec:ablation}

Our goal is to achieve optimal average results for both unoriented and oriented normal estimation.
The ablation studies in Table~\ref{tab:ablation} are discussed as follows.

\noindent\textbf{(a) Loss Functions}.
We evaluate various combinations of the proposed loss functions from Eq.~\eqref{eq:loss} to train the network model separately.
For a thorough analysis, we decompose the loss in Eq.~\eqref{eq:loss_d} as $\mathcal{L}_d \!=\! \mathcal{L}_{ld} + \mathcal{L}_{pd}$, where $\mathcal{L}_{ld} \!=\! d \!=\! \|\hat{\bm{p}} - \bm{p}\|$ and $\mathcal{L}_{pd} \!=\! d_1 + d_2$ (see Fig.~\ref{fig:dist}).
We observe that using only $\mathcal{L}_{ld}$ yields the poorest results, but the performance improves significantly when other loss terms are included.
The introduction of $\mathcal{L}_{pd}$ and $\mathcal{L}_{sd}$ is beneficial, while the addition of $\mathcal{L}_{n}$ and $\mathcal{L}_{v}$ notably enhances performance.
Some ablations yield better results in certain data categories but fail to provide consistent improvement across both unoriented and oriented normal estimation.
Our method achieves the best overall performance only when all the loss functions are applied.

\noindent\textbf{(b) Normal Aggregation}.
In Sec.~\ref{sec:normal_infer}, we propose a neighborhood weighted aggregation strategy to infer the normals of the raw data $\bm{P}$.
Here, we use the raw data as input and infer the gradient of the implicit field at each point $\bm{p} \in \bm{P}$ as its normal.
The ablation results validate the effectiveness of this inference strategy, showing improved performance in both unoriented and oriented normal estimation tasks.

\noindent\textbf{(c) Neighborhood Scale}.
For our neighborhood scale set $\{K_j\}^{N_K}_{j=1}$, we use the base parameter $\mathcal{K} \!=\! 8$ in our implementation.
Here, we test different values of $\mathcal{K}$, including $4$ and $16$.
Results indicate that while these alternative values may yield slight advantages in specific data categories, our chosen setting provides better results across both unoriented and oriented normal estimation tasks.

\section{Conclusion}

In this work, we presented a novel local gradient-aware surface filtering method for estimating oriented normals in noisy point clouds, overcoming the limitations of traditional approaches that often struggle with noise and require extensive labeled data.
By leveraging neural implicit representations and introducing specialized loss functions with local gradient constraints, our method bridges the gap between noisy point cloud data and high-quality surface representations.
Compared with the baseline methods, our method suppresses noise more effectively, achieves higher stability, and provides reliable output in different scenarios.
Experimental results across three different tasks validate the method's effectiveness and robustness, highlighting its suitability for practical 3D vision applications.
Future work may further refine this framework to extend its utility across other point cloud processing tasks.

\section*{Acknowledgements}
This work was supported by the National Natural Science Foundation of China (62402401), the Sichuan Provincial Natural Science Foundation of China (2025ZNSFSC1462) and the Fundamental Research Funds for the Central Universities (2682025CX109).

{
    \small
    \bibliographystyle{ieeenat_fullname}
    \bibliography{egbib}
}

\appendix

\section{More Experimental Results}

\begin{figure*}[t]
    \centering
    \subfloat[FamousShape dataset]{
        \includegraphics[width=.49\linewidth]{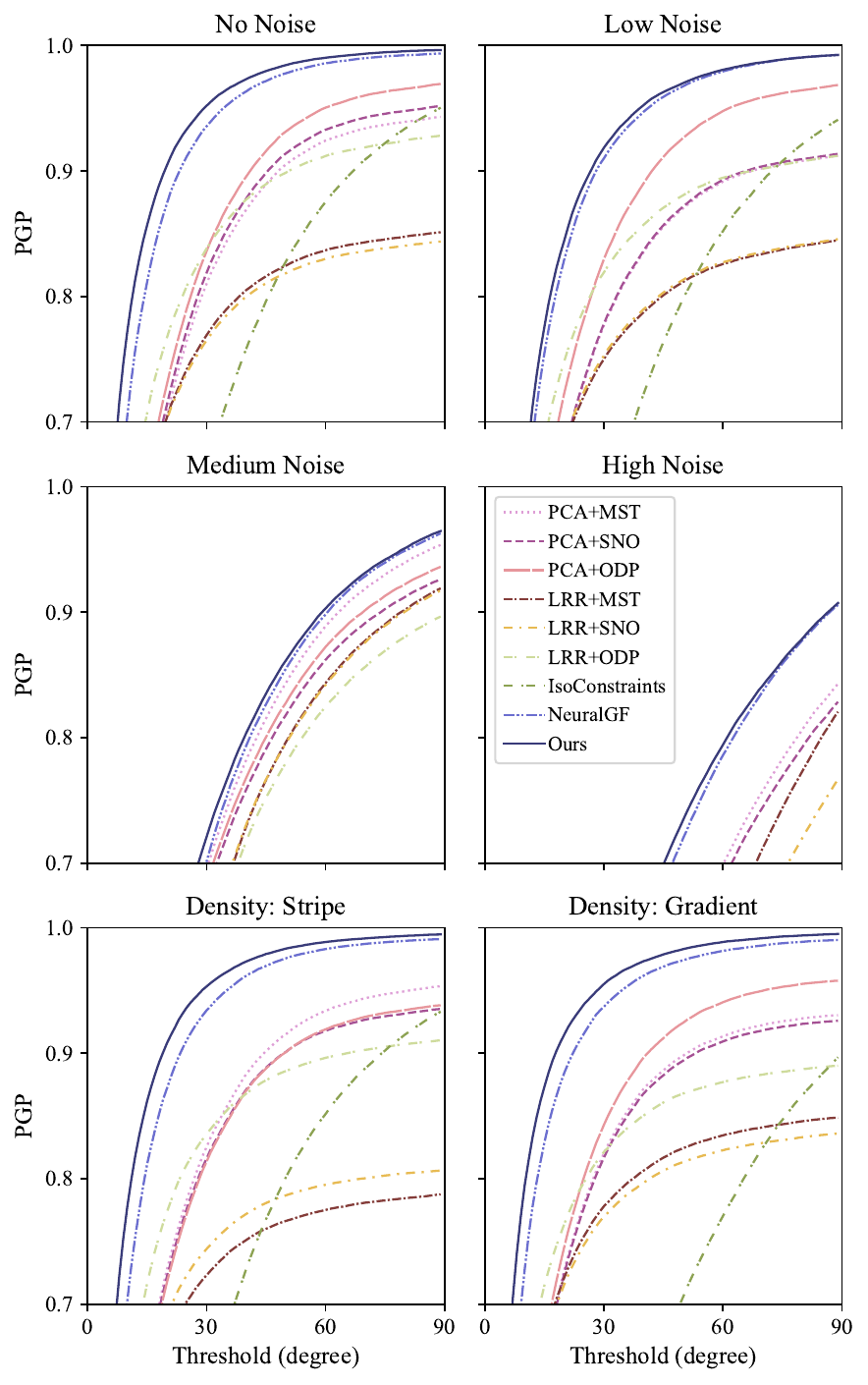}
    }
    \subfloat[PCPNet dataset]{
        \includegraphics[width=.49\linewidth]{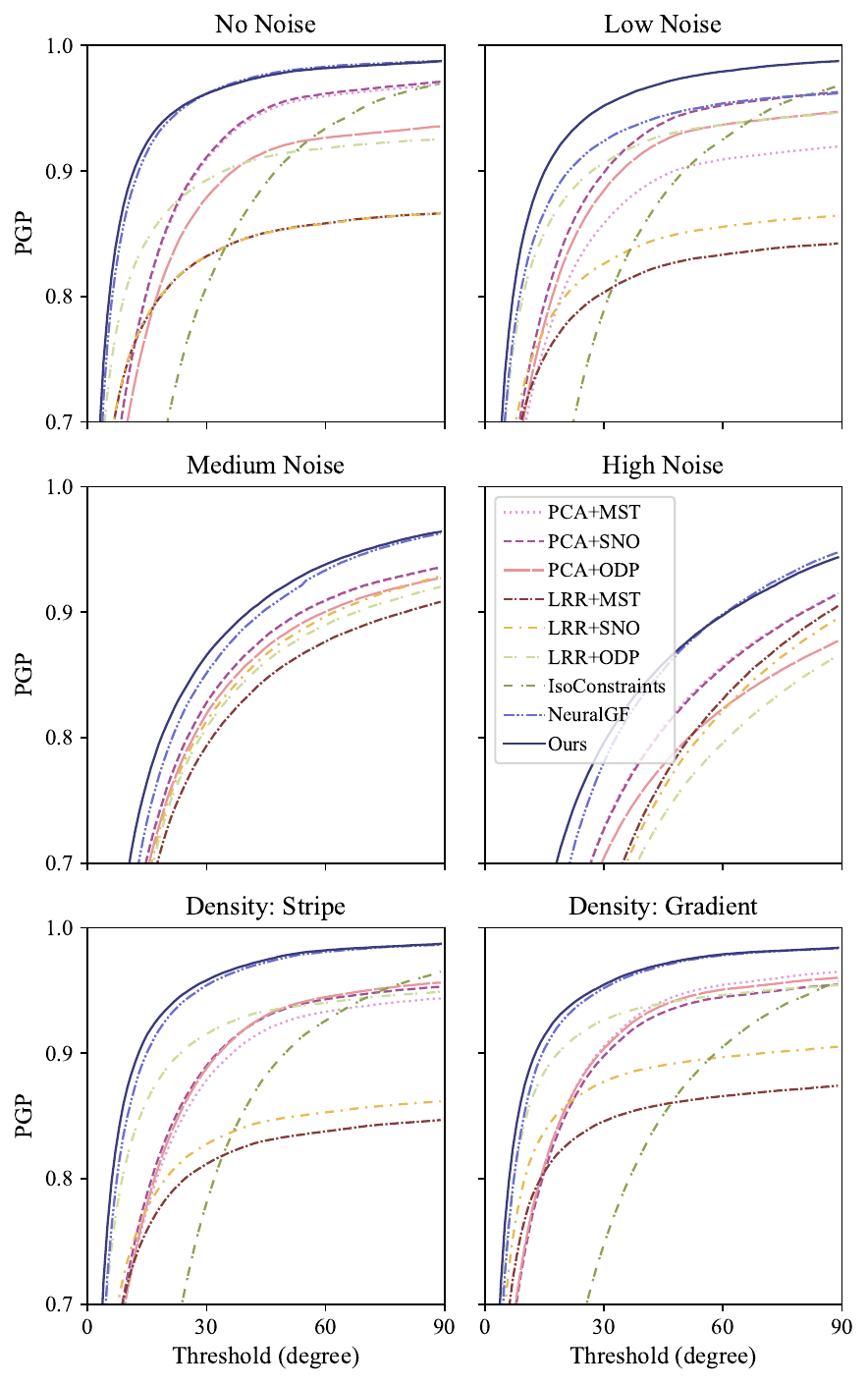}
    }
    \vspace{-0.3cm}
    \caption{
        Oriented normal PGP curves of unsupervised methods.
        The Y-axis indicates the percentage of good point normals with errors below the angle thresholds specified on the X-axis.
        Our method achieves superior results across almost all thresholds.
    }
    \label{fig:curve_FamousShape}
\end{figure*}

\begin{figure*}[t]
    \centering
    \subfloat[]{
      \includegraphics[width=0.9\linewidth]{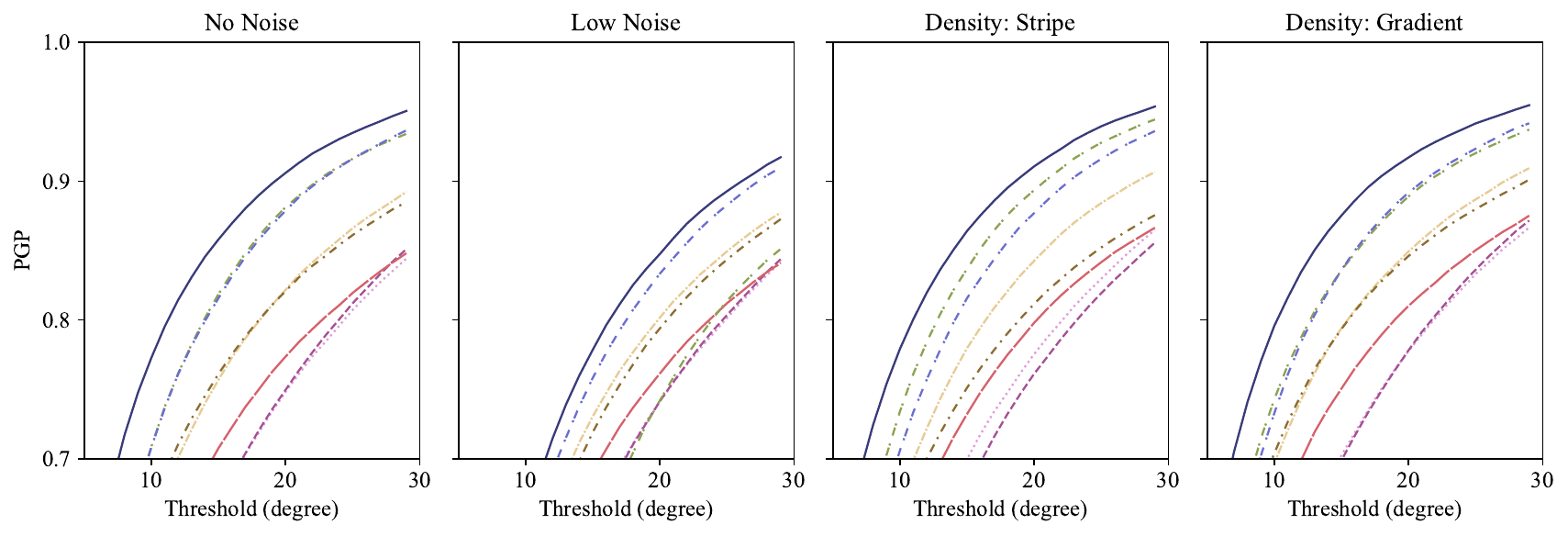}
    }
    \\
    \subfloat[]{
      \includegraphics[width=0.45\linewidth]{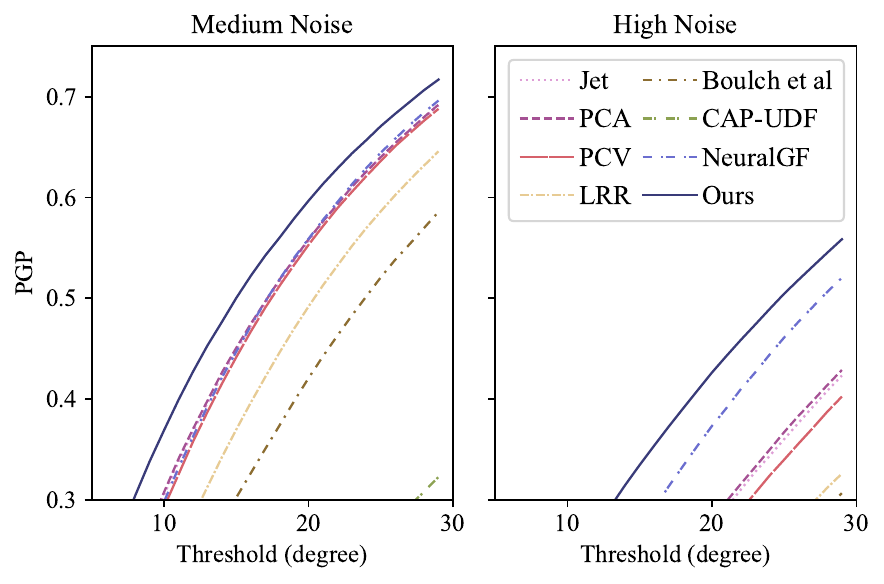}
    }
    \vspace{-0.3cm}
    \caption{
        Unoriented normal PGP curves of unsupervised methods on the FamousShape dataset.
        The Y-axis represents the percentage of good point normals whose errors are below the angle thresholds specified on the X-axis.
        The scale of the Y-axis in (a) is different from that in (b).
        Our method consistently outperforms others across all thresholds.
    }
    \label{fig:curve_uo}
\end{figure*}

\begin{figure*}[t]
    \centering
    \includegraphics[width=\linewidth]{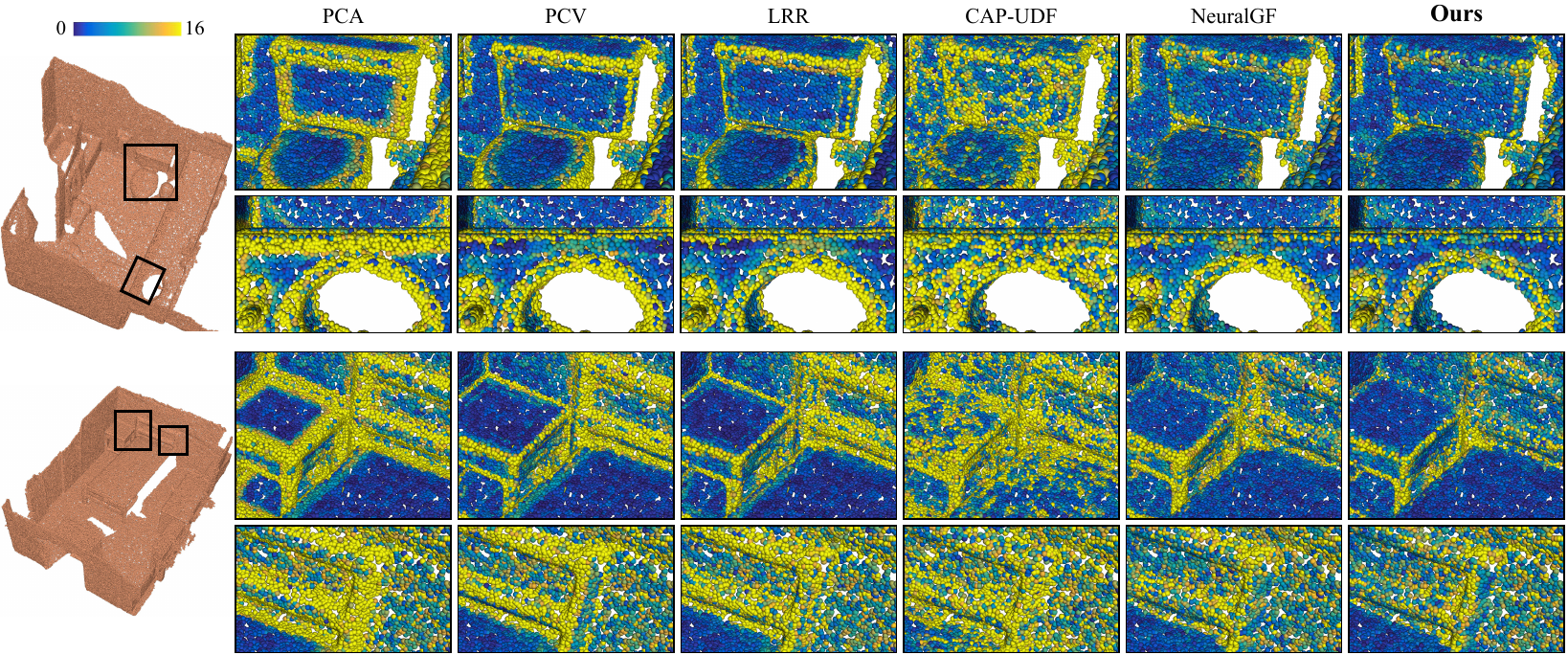}  \vspace{-0.7cm}
    \caption{
        Visual comparison of unoriented normals on real-world point clouds of the ScanNet dataset.
        Point colors indicate normal errors.
    }
    \label{fig:errorMap_ScanNet}
\end{figure*}

\noindent\textbf{Normal Estimation.}
The PGP curve represents the percentage of good point normals (PGP) under a series of given angle thresholds, providing a comprehensive evaluation of normal estimation accuracy across varying levels of precision requirements.
In Fig.~\ref{fig:curve_FamousShape}, the PGP curves of oriented normals for unsupervised methods clearly illustrate that our method consistently achieves superior performance across all thresholds and all data categories.
This highlights the robustness of our approach in capturing accurate normals, even under challenging scenarios.
Similarly, the PGP curves of unoriented normals, as depicted in Fig.~\ref{fig:curve_uo}, further demonstrate the effectiveness of our method, showcasing improved performance over baseline methods in all tested conditions.
To validate the practical applicability of our method, we extended the evaluation to real-world scanned data from the ScanNet dataset~\cite{dai2017scannet}.
This dataset provides complex, noisy, and unstructured 3D data that are captured in real-world indoor scenarios.
As shown in Fig.~\ref{fig:errorMap_ScanNet}, our approach achieves significantly improved results compared to previous methods, with visual examples demonstrating more accurate normal estimations.
Overall, the superior performance across synthetic and real-world datasets underlines the robustness, accuracy, and generalizability of our approach for normal estimation in noisy point clouds.

\begin{figure*}[t]
    \centering
    \includegraphics[width=\linewidth]{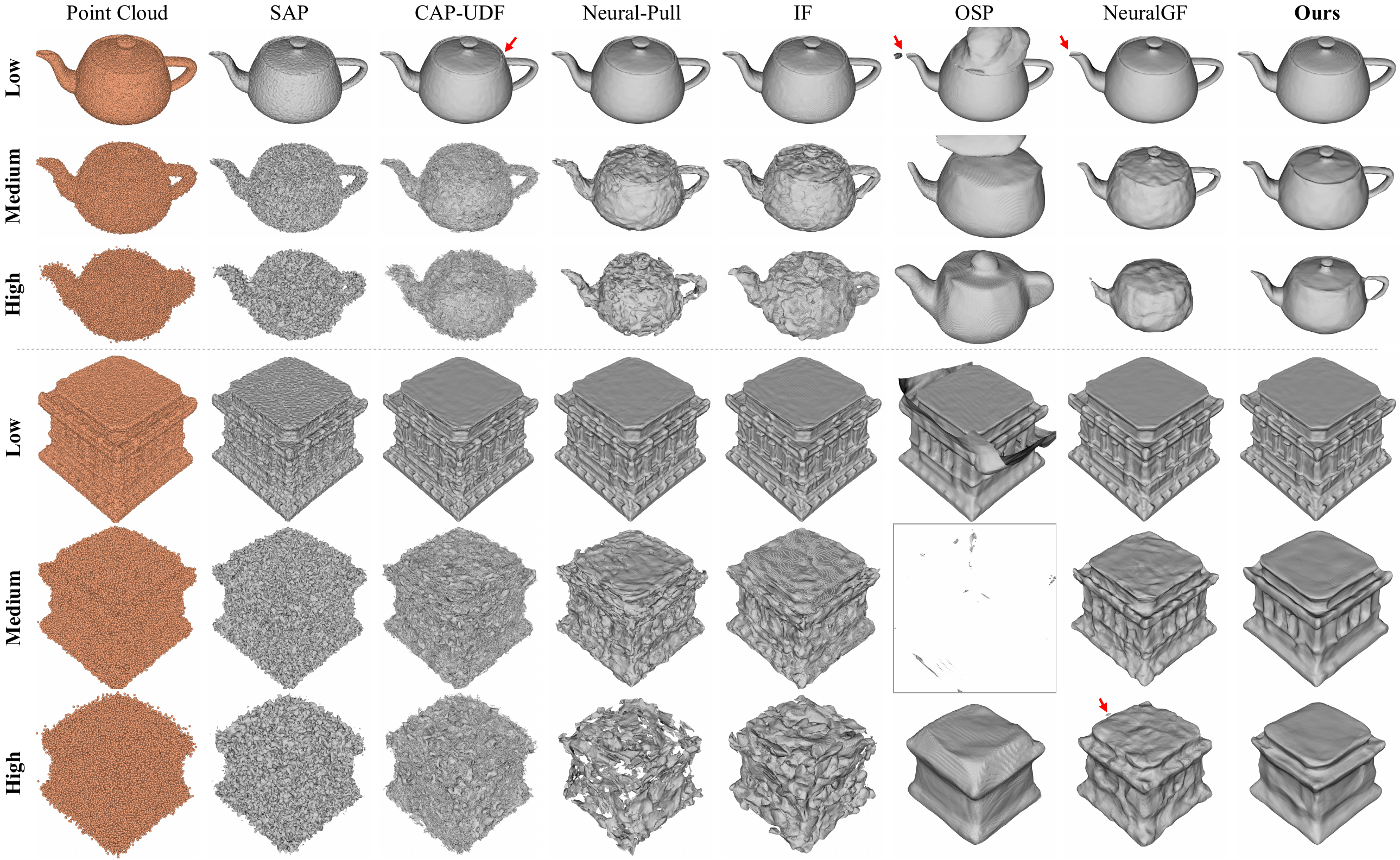}  \vspace{-0.6cm}
    \caption{
        Comparison with implicit representation methods for surface reconstruction.
        As the noise increases (from low to high), our method becomes more advantageous.
    }
    \label{fig:surfRecon_FamousShape}
\end{figure*}

\begin{figure*}[t]
    \centering
    \includegraphics[width=\linewidth]{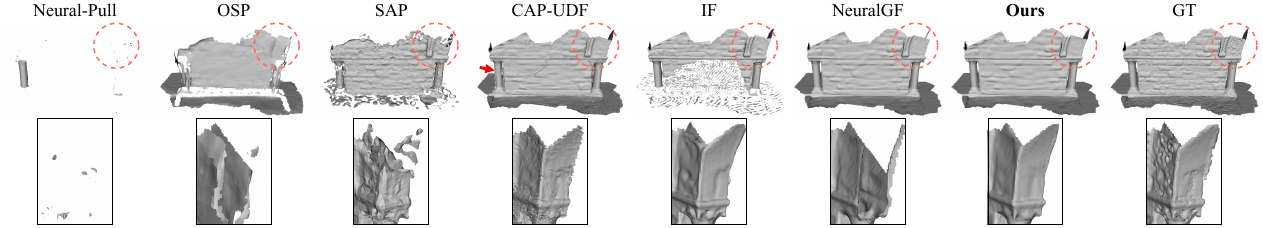}  \vspace{-0.6cm}
    \caption{
        Comparison with implicit representation methods for surface reconstruction on the 3D Scene dataset.
    }
    \label{fig:surfRecon_3DScene}
\end{figure*}

\begin{figure*}[t]
    \centering
    \includegraphics[width=.8\linewidth]{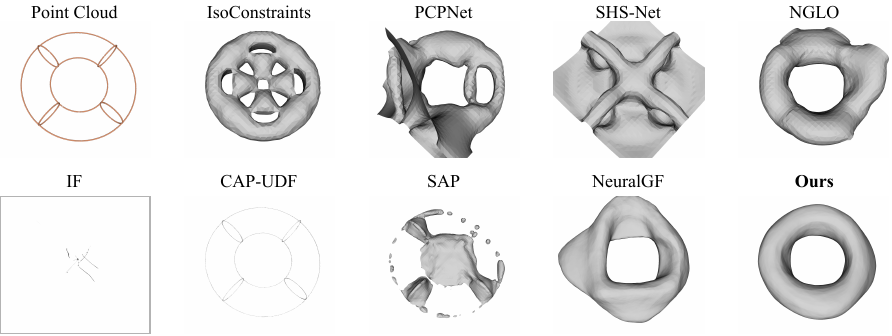}  \vspace{-0.3cm}
    \caption{
        Comparison with normal estimation methods and implicit representation methods for surface reconstruction on a wireframe point cloud of doughnut.
        The reconstructed surfaces of normal estimation methods in the first row are generated using the Poisson surface reconstruction algorithm~\cite{kazhdan2013screened}.
        The Neural-Pull~\cite{ma2020neural}, OSP~\cite{ma2022reconstructing}, IF~\cite{li2025implicit} and CAP-UDF~\cite{zhou2024cap} methods fail to generate surfaces.
    }
    \label{fig:poissonRecon_WireframePC}
    \vspace{-0.2cm}
\end{figure*}

\noindent\textbf{Surface Reconstruction.}
As illustrated in Fig.~\ref{fig:surfRecon_FamousShape}, we provide a detailed visual comparison of reconstructed surfaces generated from point clouds with varying noise levels.
These results highlight our method's ability to maintain surface integrity and recover fine details, even in the presence of significant noise, outperforming baseline approaches in terms of clarity and structural fidelity.
Further, in Fig.~\ref{fig:surfRecon_3DScene}, we visualize reconstructed surfaces from real scanned point clouds of the 3D Scene dataset~\cite{zhou2013dense}.
The results demonstrate our method's robustness in processing real-world data, effectively handling challenges such as noise, irregular sampling, and complex geometries, producing reconstructions that are both accurate and visually coherent.
In Fig.~\ref{fig:poissonRecon_WireframePC}, we explore an even more challenging scenario by generating surfaces from wireframe-like point clouds, characterized by extreme sparsity and non-uniform sampling.
Remarkably, our approach successfully reconstructs surfaces with consistent geometry and minimal artifacts, demonstrating its adaptability and resilience to highly irregular data distributions.
Overall, these visual comparisons confirm the versatility and effectiveness of our method across various challenging datasets, making it well-suited for both synthetic and real-world applications.

\begin{figure*}[t]
    \centering
    \includegraphics[width=\linewidth]{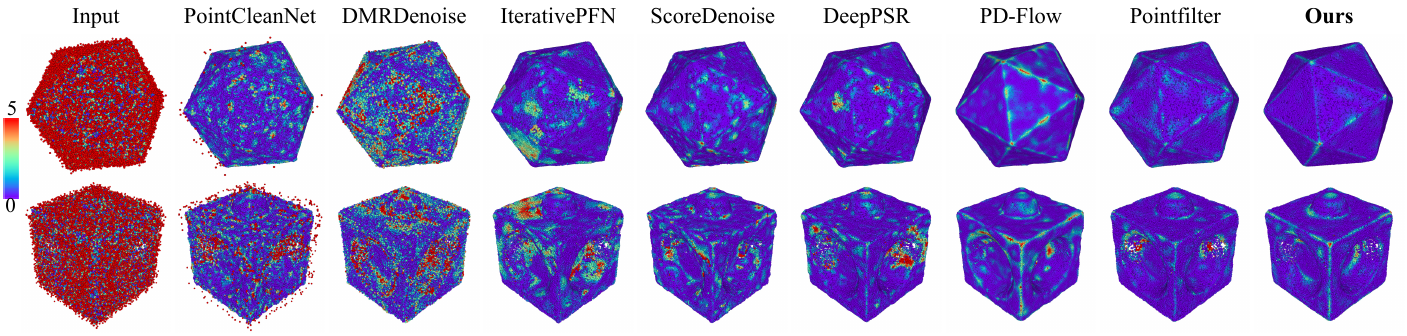}  \vspace{-0.8cm}
    \caption{
        Visual comparison of point cloud denoising.
        Our method is unsupervised, whereas all baseline methods rely on supervised training.
        The input consists of 50K-resolution shapes with 3\% Gaussian noise from the PointCleanNet dataset.
        Point colors represent P2M distance error ($\times 10^{-4}$).
    }
    \label{fig:denoise_PCNet}
    \vspace{-0.2cm}
\end{figure*}

\begin{figure}[t]
    \centering
    \vspace{-0.3cm}
    \includegraphics[width=.8\linewidth]{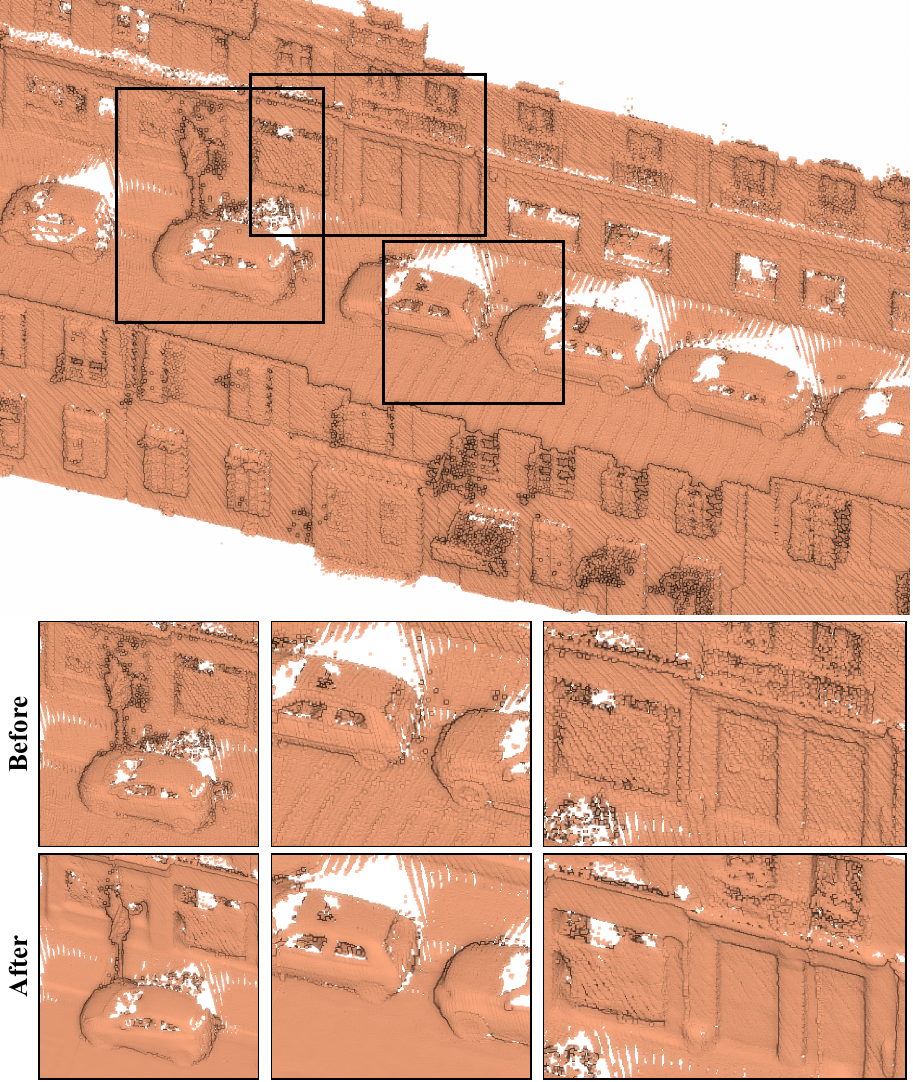}  \vspace{-0.25cm}
    \caption{
        Visualization of point cloud denoising on real-world point clouds of the Paris-rue-Madame dataset.
        We show the raw data of a street scene in the first row, and compare the local details of the scene before and after denoising using our method in the last two rows.
    }
    \label{fig:denoise_RueMadame}
    \vspace{-0.4cm}
\end{figure}

\noindent\textbf{Point Cloud Denoising.}
We compare our approach with baseline methods that rely on supervised training with ground truth labels.
In Fig.~\ref{fig:denoise_PCNet}, we present a visual comparison of denoising results on the PointCleanNet dataset~\cite{rakotosaona2020pointcleannet} for various methods.
Our approach demonstrates a superior ability to recover clean and smooth surfaces from noisy point clouds while preserving geometric details, even in challenging cases with significant noise levels.
To further evaluate its practical applicability, we also apply our method to real-world scanned data.
Fig.~\ref{fig:denoise_RueMadame} showcases denoising results on selected scenes from the Paris-Rue-Madame dataset~\cite{serna2014paris}.
Despite not requiring ground truth supervision, our method effectively mitigates noise and produces visually coherent surfaces, demonstrating its robustness and adaptability to real-world scenarios.
These results validate the effectiveness of our designed loss functions, which guide the network to accurately infer the clean surfaces from various noisy point clouds.
By leveraging statistical reasoning rather than relying on explicit ground truth, our method achieves reliable performance in both synthetic and real-world applications, setting a new standard for unsupervised point cloud denoising.
The above results confirm the suitability of our method for real-world applications in 3D vision tasks.

\noindent\textbf{Comparison with NeuralGF on runtime \& convergence.}
Since our method includes more sophisticated losses compared to NeuralGF~\cite{li2023neuralgf}, the optimization incurs a slightly higher per-instance time cost ($\sim \!+0.4\%$).
However, despite this small overhead, our method can achieve better convergence.
In Table~\ref{tab:com_gf}, we provide the comparison of average oriented-normal RMSE on the PCPNet dataset at different iterations.
The richer constraints of our method lead to lower error within the same iterations.

\begin{table}[h]
    \footnotesize
    \centering
    \setlength{\tabcolsep}{3mm}
    \caption{
        Comparison with NeuralGF on convergence.
    }
    \vspace{-0.3cm}
    \begin{tabular}{c|ccc}
        \toprule
        Iterations                                & 10,000              & 20,000                     \\
        \midrule
        NeuralGF~\cite{li2023neuralgf}            & 19.47               & 18.70                      \\
        Ours                                      & \textbf{19.22}      & \textbf{17.00}             \\
        \bottomrule
    \end{tabular}
    \label{tab:com_gf}
    \vspace{-0.2cm}
\end{table}

\section{Discussion}

\noindent\textbf{Hyperparameter tuning in loss.}
Compared to prior works, we enhance existing losses from a new perspective and introduce novel richer constraints to boost performance.
To ensure each loss term contributes meaningfully and no term is orders of magnitude larger or smaller, we let terms with larger raw values receive smaller weight factors, and vice versa.
The weights $\lambda_1$ and $\lambda_2$ are chosen via small grid searches on a held-out subset.
We find that varying each weight by $\pm 50\%$ changes the average RMSE by $<3\%$, showing low sensitivity.
For $\lambda_1$, we recommend trying $\{0.05, 0.1, 0.3\}$ to choose the value that minimizes normal RMSE.
$\lambda_2$ are less sensitive and can remain the defaults unless targeting extremely noisy or highly detailed data.
In $\mathcal{L}_{sd}$, a $10\times$ factor on the signed distances of surface points forces them to be located on the zero level set and balances surface fitting and denoising.
We will release code with default settings for transparency.

\noindent\textbf{Detail preservation and over-smoothing on real data.}
Our primary focus in this work is on accurately estimating normals from noisy point clouds.
In Fig.~\ref{fig:denoise_RueMadame}, we demonstrate an extension of our method to point cloud denoising on real-world scans from the Paris-Rue-Madame dataset.
While multiple rounds of denoising noticeably improve the visual quality of the surface, fully recovering all fine details remains challenging, particularly because real-world scans often contain a substantial number of outliers.
These outliers can skew the surface filtering process, leading to localized over-smoothing even as most noise is removed.
Our multi-round filtering on the point cloud denoising task highlights this trade-off: while most outliers and small artifacts are successfully removed, some local regions appear overly smoothed.
We acknowledge this limitation and agree that in scenes with many outliers, our method may struggle to balance outlier rejection and structure preservation.
We note that this localized over-smoothing is not unique to our approach, some state-of-the-art point cloud denoising methods exhibit similar behavior when aggressively removing noise.
To address datasets with heavy outlier contamination, a practical strategy is to incorporate a lightweight preprocessing step that detects and removes outliers before applying the denoising pipeline.
Integrating robust outlier filtering or adaptive smoothing strategy into the end-to-end framework is an important direction for future work.

\section{Limitation}

Our method can be applied to various point cloud processing tasks, as demonstrated in our experiments.
However, a limitation of our approach is that the neural network must be optimized individually for each point cloud.
As a result, the trained model cannot be directly applied to shapes that were not part of the optimization process.
This limitation is similar to some implicit representation methods~\cite{ma2020neural,ma2022reconstructing,zhou2024cap,li2025implicit} that overfit to individual data, making our method less generalizable compared to pre-trained models like SHS-Net~\cite{li2024shsnet-pami} that can be used out-of-the-box for normal estimation on new point cloud data.
Therefore, it will incur additional time costs for users to directly start using the proposed method for normal estimation.
Future work will focus on addressing this limitation by generalizing the method to handle unseen data, enabling more convenient and broader applicability.

\noindent\textbf{Potential directions for improving the generalization.}
(1) Front-end feature extractor: mapping objects to a unified metric space using shape-specific features, enabling fast adaptation to diverse geometries.
(2) Meta-learning or model-agnostic initialization: pretraining on a small corpus of shapes to warm-start the network for unseen objects.



\end{document}